\documentclass[nohyperref]{article}

\usepackage{microtype}
\usepackage{graphicx}
\usepackage{booktabs} 

\usepackage{hyperref}

\usepackage[accepted]{icml2022}

\usepackage{amsmath}
\usepackage{amssymb}
\usepackage{mathtools}
\usepackage{amsthm}
\usepackage{times}
\usepackage{latexsym}
\usepackage{graphicx}
\usepackage{bm}
\usepackage{multirow}
\usepackage{subcaption, booktabs} 
\usepackage[T1]{fontenc}
\usepackage{soulutf8}
\usepackage{url}
\usepackage{tablefootnote}

\usepackage[capitalize,noabbrev]{cleveref}

\theoremstyle{plain}

\theoremstyle{definition}

\theoremstyle{remark}

\usepackage{xcolor}
\usepackage{soul}
\newcommand{\hlc}[2][yellow]{{%
    \colorlet{foo}{#1}%
    \sethlcolor{foo}\hl{#2}}%
}

\definecolor{red}{RGB}{255, 22, 22}
\definecolor{blue}{RGB}{22, 22, 255}

\usepackage[textsize=tiny]{todonotes}

\icmltitlerunning{Interpreting Language Models with Contrastive Explanations}

\begin{document}
\twocolumn[
\icmltitle{Interpreting Language Models with Contrastive Explanations}

\icmlsetsymbol{equal}{*}

\begin{icmlauthorlist}
\icmlauthor{Kayo Yin}{cmu}
\icmlauthor{Graham Neubig}{cmu}
\end{icmlauthorlist}

  \author{Kayo Yin, Graham Neubig\\
  Language Technologies Institute, Carnegie Mellon University \\
  \texttt{\{kayoy, gneubig\}@cs.cmu.edu} \\}

\icmlaffiliation{cmu}{Language Technologies Institute, Carnegie Mellon University}

\icmlcorrespondingauthor{Kayo Yin}{kayoy@cs.cmu.edu}
\icmlcorrespondingauthor{Graham Neubig}{gneubig@cs.cmu.edu}

\icmlkeywords{Model Interpretability, Natural Language Processing}

\vskip 0.3in
]


\printAffiliationsAndNotice{}  

\begin{abstract}
Model interpretability methods are often used to explain NLP model decisions on tasks such as text classification, where the output space is relatively small. However, when applied to language generation, where the output space often consists of tens of thousands of tokens, these methods are unable to provide informative explanations. 
Language models must consider various features to predict a token, such as its part of speech, number, tense, or semantics.
Existing explanation methods conflate evidence for all these features into a single explanation, which is less interpretable for human understanding.

To disentangle the different decisions in language modeling, we focus on explaining language models \textit{contrastively}: we look for salient input tokens that explain why the model predicted one token \textit{instead of} another. We demonstrate that contrastive explanations are quantifiably better than non-contrastive explanations in verifying major grammatical phenomena, and that they significantly improve contrastive model simulatability for human observers. We also identify groups of contrastive decisions where the model uses similar evidence, and we are able to characterize what input tokens models use during various language generation decisions.\footnote{Code and demo at \url{https://github.com/kayoyin/interpret-lm}}
\end{abstract}

\section{Introduction}

Despite their success across a wide swath of natural language processing (NLP) tasks, neural language models (LMs) are often used as black boxes, where \textit{how} they make certain predictions remains obscure \cite{belinkov-glass-2019-analysis} 
This is in part due to the high complexity of the task of language modeling itself, as well as that of the model architectures used to solve it.

In this paper, we argue that this is also due to the fact that interpretability methods commonly used for other NLP tasks like text classification, such as gradient-based saliency maps \cite{li-etal-2016-visualizing, sundararajan2017axiomatic}, are not as informative for LM predictions.
For example, to explain why an LM predicts \textit{``barking''} given ``Can you stop the dog from \_\_\_\_'', we demonstrate in experiments that the input token preceding the prediction is often marked as the most influential token to the prediction (Table \ref{table:main})
by instance attribution methods such as gradient $\times$ input \cite{baehrens2010explain}. The preceding token is indeed highly important to determine \emph{certain} features of the next token, ruling out words that would obviously violate syntax in that context such as predicting a verb without ``-ing'' in the given example.
However, this does not explain why the model made other more subtle decisions, such as why it predicts \textit{``barking''} instead of \textit{``crying''} or \textit{``walking''}, which are all plausible choices if we only look at the preceding token. In general, language modeling has a large output space and a high complexity compared to other NLP tasks; at each time step, the LM chooses one word out of all vocabulary items. This contrasts with text classification, for example, where the output space is smaller, because several linguistic distinctions come into play for each language model decision. 

\begin{table}[t]
\centering
\resizebox{\linewidth}{!}{ 
\begin{tabular}{l} \toprule
\textbf{Input:} \textit{\textit{Can you stop the dog from}} \\
\textbf{Output:} barking \\
\midrule
\textbf{1. Why did the model predict ``barking''?} \\
\hlc[red!46]{Can} \hlc[blue!12]{you} \hlc[blue!16]{stop} \hlc[blue!32]{the} \hlc[blue!4]{dog} \hlc[red!88]{from}\\
\midrule
\textbf{2. Why did the model predict ``barking'' \textit{instead of} ``crying''?} \\
\hlc[red!40]{Can} \hlc[blue!50]{you} \hlc[red!26]{stop} \hlc[blue!36]{the} \hlc[red!46]{dog} \hlc[blue!4]{from}\\
\midrule
\textbf{3. Why did the model predict ``barking'' \textit{instead of} ``walking''?} \\
\hlc[red!14]{Can} \hlc[blue!46]{you} \hlc[red!42]{stop} \hlc[blue!66]{the} \hlc[red!22]{dog} \hlc[red!0]{from}\\
\bottomrule
\end{tabular}}
\caption{Explanations for the GPT-2 language model prediction given the input ``Can you stop the dog from \_\_\_\_\_". Input tokens that are measured to raise or lower the probability of ``barking'' are in red and blue respectively, and those with little influence are in white. Non-contrastive explanations such as gradient $\times$ input (1) usually attribute the highest saliency to the token immediately preceding the prediction. Contrastive explanations (2, 3) give a more fine-grained and informative explanation on why the model predicted one token over another.}
\label{table:main}
\end{table}

  
  
  

To better explain LM decisions, we propose interpreting LMs with \textit{contrastive explanations} \cite{lipton_1990}. Contrastive explanations aim to identify causal factors that lead the model to produce one output \textbf{instead of} another output. 
We believe that contrastive explanations are especially useful to handle the complexity and the large output space of language modeling, as well as related tasks like machine translation (Appendix \ref{sec:nmt}). In Table \ref{table:main}, the second explanation suggests that the input word ``dog'' makes ``barking'' more likely than a verb not typical for dogs such as ``crying'', and the third explanation suggests that the input word ``stop'' increases the likelihood of ``barking'' over a verb without negative connotations such as ``walking''.


In this paper, first, we describe how we extended three existing interpretability methods to provide contrastive explanations (\S\ref{sec:methods}). We then perform a battery of experiments aimed at examining to what extent these contrastive explanations are superior to their non-contrastive counterparts from a variety of perspectives:
\begin{itemize}
    \item RQ1: Are contrastive explanations better at identifying evidence that we believe, \textit{a-priori}, to be useful to capture a variety of linguistic phenomena (\S\ref{sec:blimp-eval})?
    \item RQ2: Do contrastive explanations allow human observers to better simulate language model behavior (\S\ref{sec:human})?
    \item RQ3: Are different types of evidence necessary to disambiguate different types of words, and does the evidence needed reflect (or uncover) coherent linguistic concepts (\S\ref{sec:cluster})?
\end{itemize}

\section{Background}

\subsection{Model Explanation}

Our work focuses on model explanations that communicate \textit{why} a computational model made a certain prediction. Particularly, we focus on methods that compute \textbf{saliency scores} $S(x_i)$ over input features $x_i$ to reveal which input tokens are most relevant for a prediction: the higher the saliency score, the more $x_i$ supposedly contributed to the model output.

Despite a reasonably large body of literature examining input feature explanations for NLP models on tasks such as text classification (for a complete review see \citet{belinkov-glass-2019-analysis, madsen2021post}), there are only a few works that attempt to explain language modeling predictions, for example by applying existing non-contrastive explanation methods to BERT's masked language modeling objective \cite{Wallace2019AllenNLP}.
Despite the importance of both language models and interpretability in the NLP literature, the relative paucity of work in this area may be somewhat surprising, and we posit that this may be due to the large output space of language models necessitating the use of techniques such as contrastive explanations, which we detail further below. 


\subsection{Contrastive Explanations}

\textbf{Contrastive} explanations attempt to explain \textit{why} given an input $x$ the model predicts a \textbf{target} $y_t$ \textit{instead of} a \textbf{foil} $y_f$. Relatedly, \textbf{counterfactual} explanations explore \textit{how} to modify the input $x$ so that the model more likely predicts $y_f$ instead of $y_t$ \cite{mcgill1993contrastive}.

While contrastive and counterfactual explanations have been explored to interpret model decisions (see \citet{stepin2021survey} for a broad survey), they are relatively new to NLP and have not yet been studied to explain language models.


Recently, \citet{jacovi-etal-2021-contrastive} produce counterfactual explanations for text classification models by erasing certain features from the input, then projecting the input representation to the ``contrastive space'' that minimally separates two decision classes. Then, they measure the importance of the intervened factor by comparing model probabilities for the two contrastive classes before and after the intervention.

We, on the other hand, propose contrastive explanations for language modeling, where both the number of input factors and the output space are much larger. While we also use a counterfactual approach with input token erasure (\S\ref{sec:erasure}), counterfactual methods may become intractable over long input sequences and a large foil space. We, therefore, also propose contrastive explanations using gradient-based methods (\S\ref{sec:norm},\S\ref{sec:input}) that measure the saliency of input tokens for a contrastive model decision.



\section{Contrastive Explanations for Language Models}
\label{sec:methods}
In this section, we describe how we extend three existing input saliency methods to the contrastive setting. These methods can also be easily adapted to tasks beyond language modeling, such as machine translation (Appendix \ref{sec:nmt}).

\subsection{Gradient Norm}
\label{sec:norm}

\citet{simonyan2013deep, li-etal-2016-visualizing} calculate saliency scores based on the norm of the gradient of the model prediction, such as the output logit,  with respect to the input.
Applying this method to LMs entails first calculating the gradient as follows:
$$
g(x_i) = \nabla_{x_i} q(y_t | \bm{x})
$$
where $\bm{x}$ is the input sequence embedding, $y_t$ is the next token in the input sequence, $q(y_t | \bm{x})$ is the model output for the token $y_t$ given the input $\bm{x}$.

Then, we obtain the saliency score for the input token $x_i$ by taking the L1 norm:
$$
S_{GN}(x_i) = ||g(x_i)||_{L1}
$$

We extend this method to the \textbf{Contrastive Gradient Norm} defined by:
$$
g^*(x_i) = \nabla_{x_i} \left( q(y_t | \bm{x}) - q(y_f | \bm{x}) \right)
$$ $$
S_{GN}^*(x_i) = ||g^*(x_i)||_{L1}
$$
where $q(y_f | \bm{x})$ is the model output for foil token $y_f$ given the input $\bm{x}$. This tells us how much an input token $x_i$ influences the model to increase the probability of $y_t$ while decreasing the probability of $y_f$.

\subsection{Gradient $\times$ Input}
\label{sec:input}

For the gradient $\times$ input method \cite{shrikumar2016not, denil2014extraction}, instead of taking the L1 norm of the gradient, we take the dot product of the gradient with the input token embedding $x_i$:
$$
S_{GI}(x_i) = g(x_i) \cdot x_i
$$
By multiplying the gradient with the input embedding, we also account for how much each token is expressed in the saliency score.

We define the \textbf{Contrastive Gradient $\times$ Input} as:
$$
S^*_{GI}(x_i) = g^*(x_i) \cdot x_i
$$

\begin{table*}[htp!]
\centering
\resizebox{\linewidth}{!}{ 
\begin{tabular}{ccllc}
\toprule
Phenomenon & UID$^3$  & Acceptable Example & Unacceptable Example  & Rule\\ 
\midrule
\multirow{2}{*}{Anaphor Agreement} & aga & \ul{Katherine} can’t help \textbf{herself}. & \ul{Katherine} can’t help \textbf{himself}. & \texttt{coref} \\
& ana & Many \ul{teenagers} were helping \textbf{themselves}. & Many \ul{teenagers} were helping \textbf{herself}. & \texttt{coref} \\
\midrule
Argument Structure & asp & Amanda was \ul{respected} by some \textbf{waitresses}. & Amanda was \ul{respected} by some \textbf{picture}. & \texttt{main\_verb}\\
\midrule
\multirow{4}{*}{Determiner-Noun Agreement} & dna & Craig explored \ul{that} grocery \textbf{store}.& Craig explored \ul{that} grocery \textbf{stores}. & \texttt{det\_noun} \\
& dnai & Phillip was lifting \ul{this} \textbf{mouse}. & Phillip was lifting \ul{this} \textbf{mice}. & \texttt{det\_noun} \\
& dnaa & Tracy praises \ul{those} lucky \textbf{guys}. & Tracy praises \ul{those} lucky \textbf{guy}. & \texttt{det\_noun} \\
& dnaai & This person shouldn’t criticize \ul{this} upset \textbf{child}.& This person shouldn’t criticize \ul{this} upset \textbf{children}. & \texttt{det\_noun} \\
\midrule
NPI Licensing & npi & \ul{Even} these trucks have \textbf{often} slowed. & \ul{Even} these trucks have \textbf{ever} slowed. & \texttt{npi} \\
\midrule
\multirow{3}{*}{Subject-Verb Agreement} & darn & \ul{A sketch of lights} \textbf{doesn’t} appear. & \ul{A sketch of lights} \textbf{don’t} appear. & \texttt{subj\_verb} \\
& ipsv & This \ul{goose} \textbf{isn’t} bothering Edward. & This \ul{goose} \textbf{weren’t} bothering Edward. & \texttt{subj\_verb} \\
& rpsv & \ul{Jeffrey} \textbf{hasn’t} criticized Donald. & \ul{Jeffrey} \textbf{haven’t} criticized Donald. & \texttt{subj\_verb} \\
\bottomrule
\end{tabular}}
\caption{Examples of minimal pairs in BLiMP. Contrastive tokens are \textbf{bolded}. Important tokens extracted by our rules are \ul{underlined}.} 
\label{table:blimp}
\end{table*}

\subsection{Input Erasure}
\label{sec:erasure}
Erasure-based methods measure how erasing different parts of the input affects the model output \cite{li2016understanding}. This can be done by taking the difference between the model output with the full input $\bm{x}$ and with the input where $x_i$ has been zeroed out, $\bm{x}_{\neg i}$:
$$
S_{E}(x_i) = q(y_t | \bm{x}) - q(y_t | \bm{x}_{\neg i})
$$

We define the \textbf{Contrastive Input Erasure} as:
$$
S^*_{E}(x_i) = \left(q(y_t | \bm{x}) - q(y_t | \bm{x}_{\neg i})\right) - \left(q(y_f | \bm{x}) - q(y_f | \bm{x}_{\neg i})\right)
$$
This measures how much erasing $x_i$ from the input makes the foil more likely and the target less likely in the model output. 

Although erasure-based methods directly measure the change in the output due to a perturbation in the input, while gradient-based methods approximate this measurement, erasure is usually more computationally expensive due to having to run the model on all possible input perturbations.

\section{Do Contrastive Explanations Identify Linguistically Appropriate Evidence?}
\label{sec:blimp-eval}
First, we ask whether contrastive explanations are quantifiably better than non-contrastive explanations in identifying evidence that we believe \textit{a priori} should be important to the LM decision. In order to do so, we develop a methodology in which we specify certain types of evidence that should indicate how to make particular types of linguistic distinctions, and measure how well each variety of explanation method uncovers this specified evidence.

\subsection{Linguistic Phenomena}
As a source of linguistic phenomena to study, we use the BLiMP dataset \cite{warstadt-etal-2020-blimp-benchmark}. This dataset contains 67 sets of 1,000 pairs of minimally different English sentences that contrast in grammatical acceptability under a certain linguistic paradigm. An example of a linguistic paradigm may be \textit{anaphor number agreement}, where an acceptable sentence is ``Many teenagers were helping \textbf{themselves}.'' and a minimally contrastive unacceptable sentence is ``Many teenagers were helping \textbf{herself}.'' because in the latter, the number of the reflexive pronoun does not agree with its antecedent.

From this dataset, we chose 5 linguistic phenomena with 12 paradigms and created a set of rules for each phenomenon to identify the input tokens that enforce grammatical acceptability. In the previous example, the anaphor number agreement is enforced by the antecedent ``teenagers''. We show an example of each paradigm and its associated rule in Table \ref{table:blimp} and we describe them in the following. Note that these rules are designed to be used on the BLiMP dataset and may not always generalize to other data.

\paragraph{Anaphor Agreement.} The gender and the number of a pronoun must agree with its antecedent. We implement the \texttt{coref} rule using spaCy \cite{spacy2} and NeuralCoref\footnote{\url{https://github.com/huggingface/neuralcoref}} to extract all input tokens that are coreferent with the target token.

\paragraph{Argument Structure.} Certain arguments can only appear with certain types of verbs. For example, action verbs must often be used with animate objects. We implement the \texttt{main\_verb} rule using spaCy to extract the main verb of the input sentence.

\paragraph{Determiner-Noun Agreement.} Demonstrative determiners and the associated noun must agree. We implement the \texttt{det\_noun} rule by generating the dependency tree using spaCy and extracting the determiner of the target noun.

\paragraph{NPI Licensing.} Certain negative polarity items (NPI) are only allowed to appear in certain contexts, e.g.~``never'' appears on its own in sentences, while the word ``ever'' generally must be preceded by ``not''. In all of our examples with NPI licensing, the word ``even'' is an NPI that can appear in the acceptable example but not in the unacceptable example, so we create the \texttt{npi} rule that extracts this NPI.

\paragraph{Subject-Verb Agreement.} The number of the subject and its verb in the present tense must agree. We implement the \texttt{subj\_verb} rule by generating the dependency tree using spaCy and extracting the subject of the target verb.

\addtocounter{footnote}{1}
\footnotetext{BLiMP Unique Identifiers for paradigms. aga: anaphor\_gender\_agreement; ana: anaphor\_number\_agreement; asp: animate\_subject\_passive; dna: determiner\_noun\_agreement\_1; dnai: determiner\_noun\_agreement\_irregular\_1; dnaa: determiner\_noun\_agreement\_with\_adj\_1; dnaai: determiner\_noun\_agreement\_with\_adj\_irregular\_1; npi: npi\_present\_1; darn: distractor\_agreement\_relational\_noun; ipsv: irregular\_plural\_subject\_verb\_agreement\_1; rpsv: regular\_plural\_subject\_verb\_agreement\_1}

\subsection{Alignment Metrics}
We use three metrics to quantify the alignment between an explanation and the known evidence enforcing a linguistic paradigm. The explanation is a vector $\bm{\mathcal{S}}$ of the same size as the input $\bm{x}$, where the $i$-th element $\mathcal{S}_i$ gives the saliency score of the input token $x_i$. The known evidence is represented with a binary vector $\bm{\mathcal{E}}$, also of same size as the input $\bm{x}$, where $\mathcal{E}_i = 1$ if the token $x_i$ enforces a grammatical rule on the model decision.

\paragraph{Dot Product.}
The dot product $\bm{\mathcal{S}} \cdot \bm{\mathcal{E}}$ measures the sum of saliency scores of all input tokens that are part of the known evidence. 

\paragraph{Probes Needed \cite{faithful-attn, yin-etal-2021-context}.}  We measure the number of tokens we need to probe, based on the explanation $\bm{\mathcal{S}}$, to find a token that is in the known evidence. This corresponds to the ranking of the first token $x_i$ such that $G_i = 1$ after sorting all tokens by descending saliency. 

\paragraph{Mean Reciprocal Rank (MRR).}
We calculate the average of the inverse of the rank of the first token that is part of the known evidence if the tokens are sorted in descending saliency. This also corresponds to the average of the inverse of the probes needed for each sentence evaluated.

``Dot Product'' and ``Probes Needed'' calculate the alignment for each sentence, and we compute the average over all sentence-wise alignment scores for the alignment score over a linguistic paradigm. ``MRR'' directly calculates the alignment over an entire paradigm.

\begin{figure}[htp!]
\centering
\begin{minipage}{\linewidth}
\centering
\includegraphics[height=2.85cm]{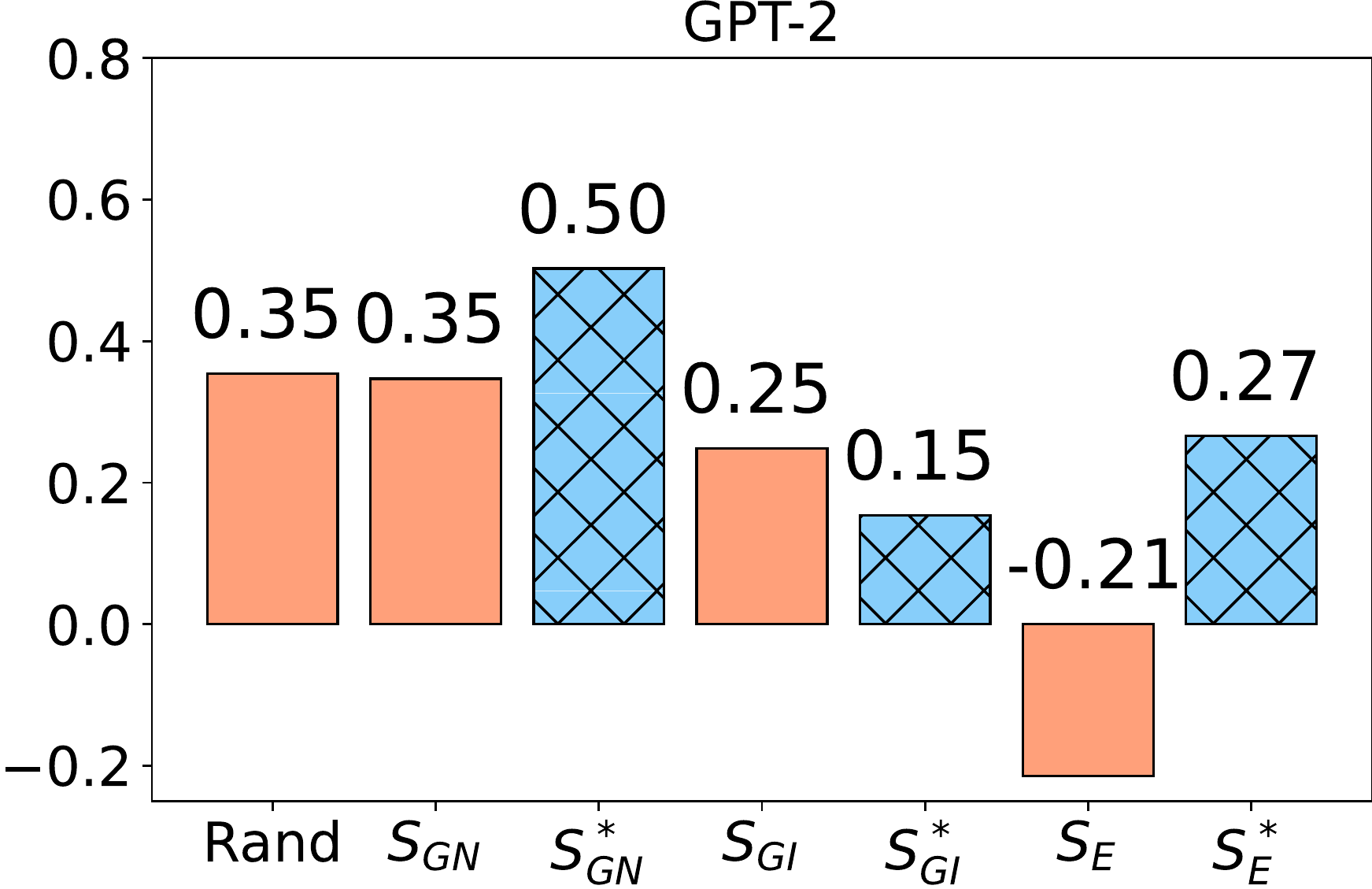}%
\includegraphics[height=2.85cm]{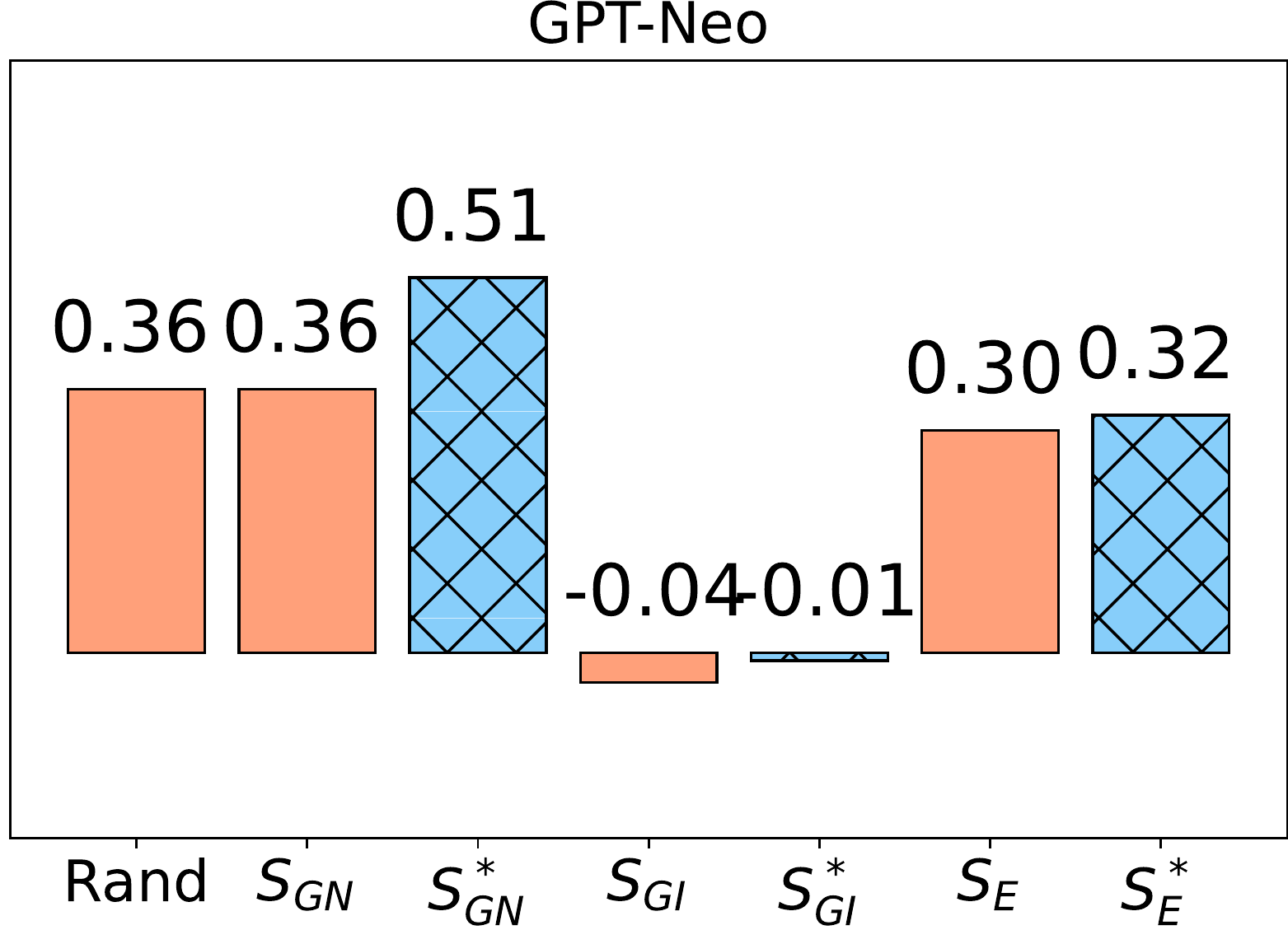}%
\subcaption{Dot Product (↑)}
\end{minipage}
\begin{minipage}{\linewidth}
\includegraphics[height=2.9cm]{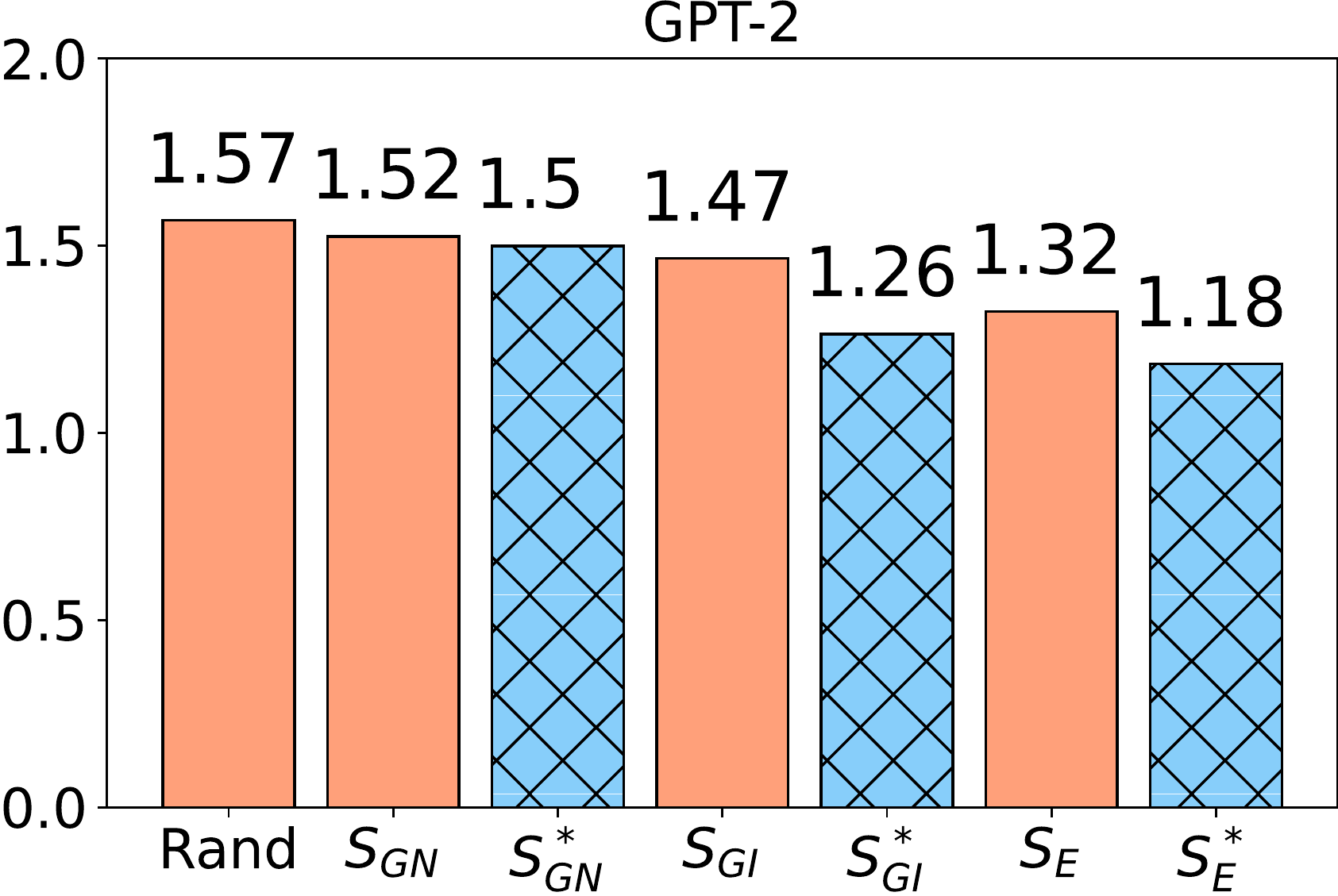}%
\includegraphics[height=2.9cm]{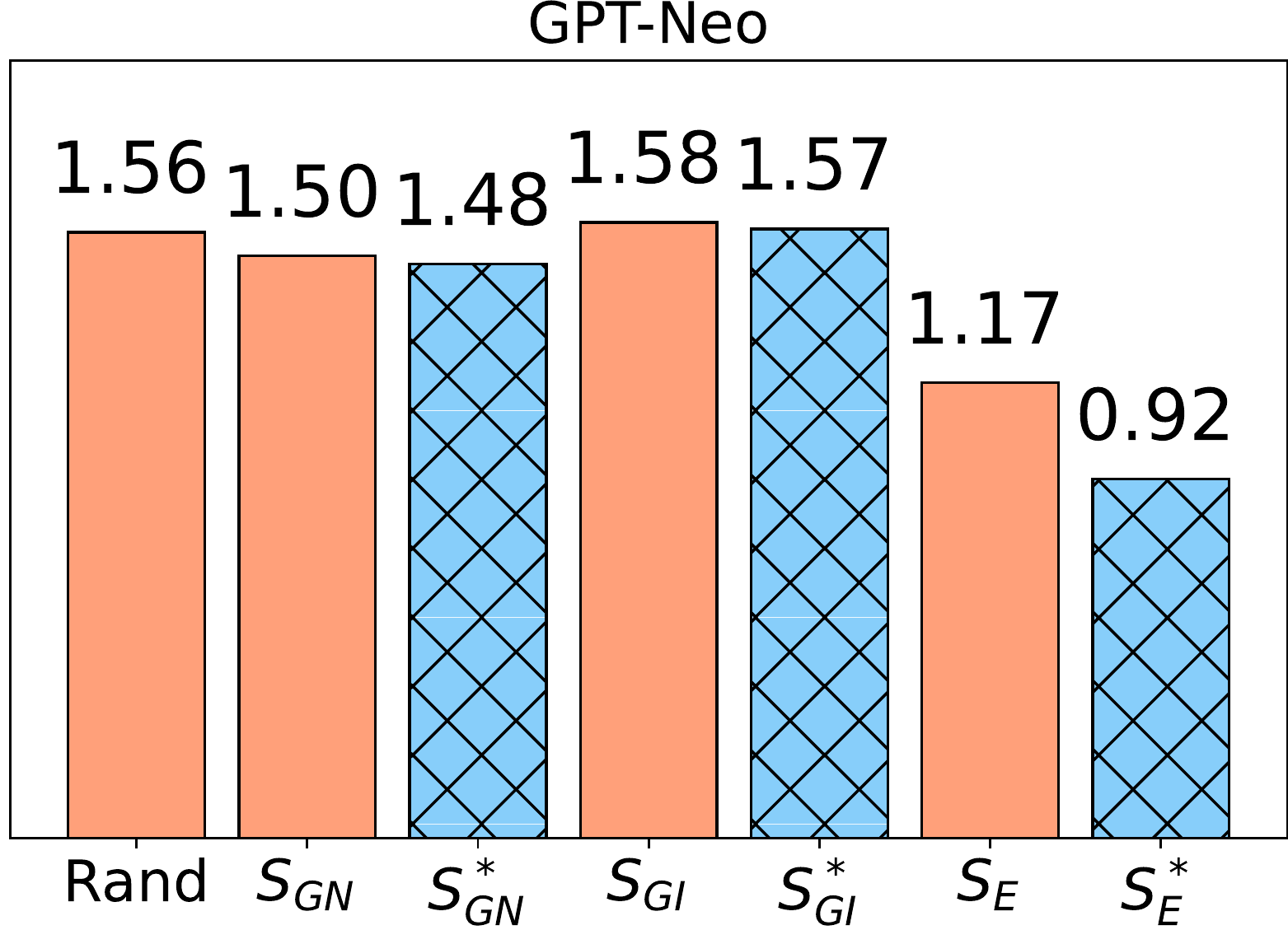}%
\subcaption{Probes Needed (↓)}
\end{minipage}
\begin{minipage}{\linewidth}
\includegraphics[height=2.9cm]{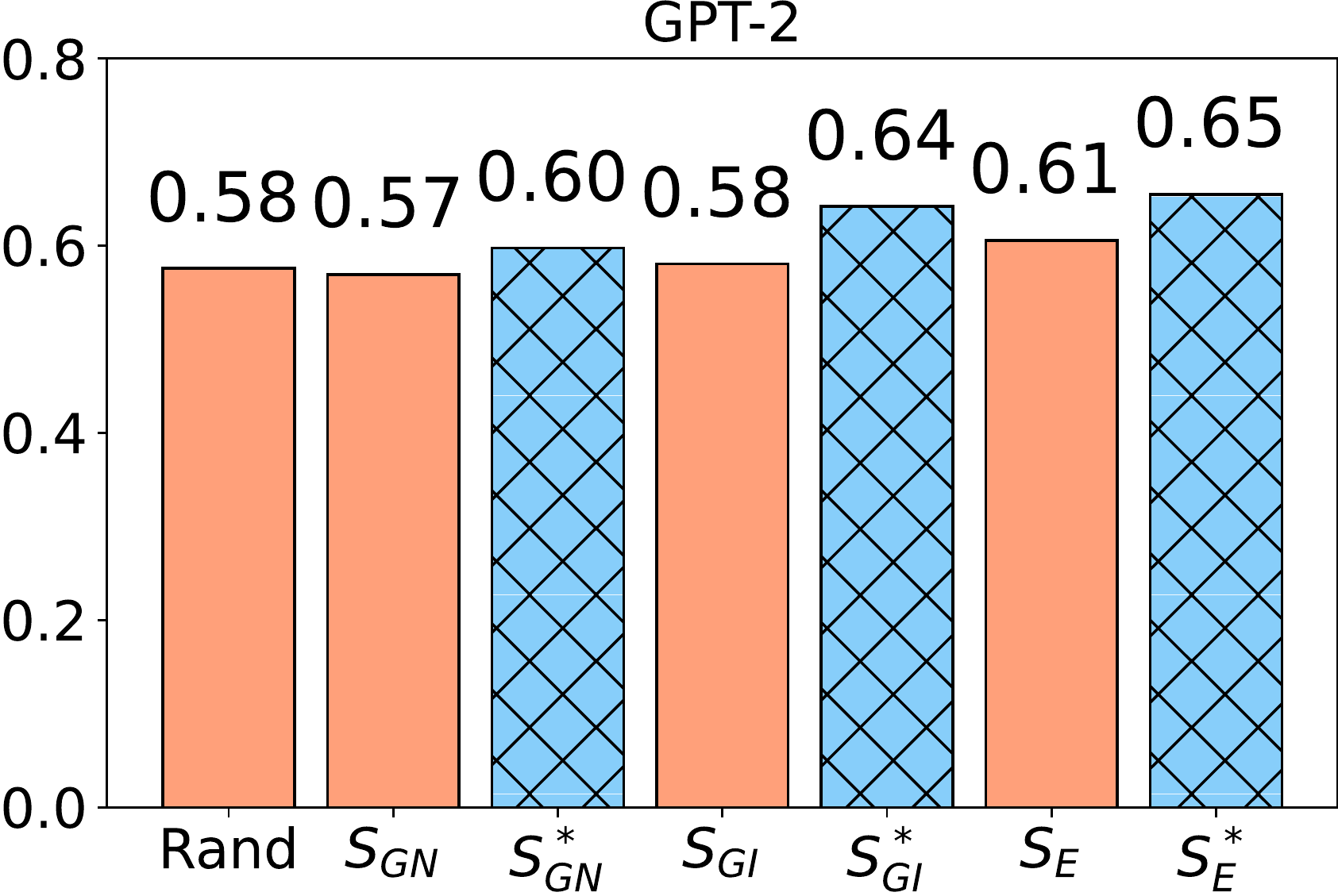}%
\includegraphics[height=2.9cm]{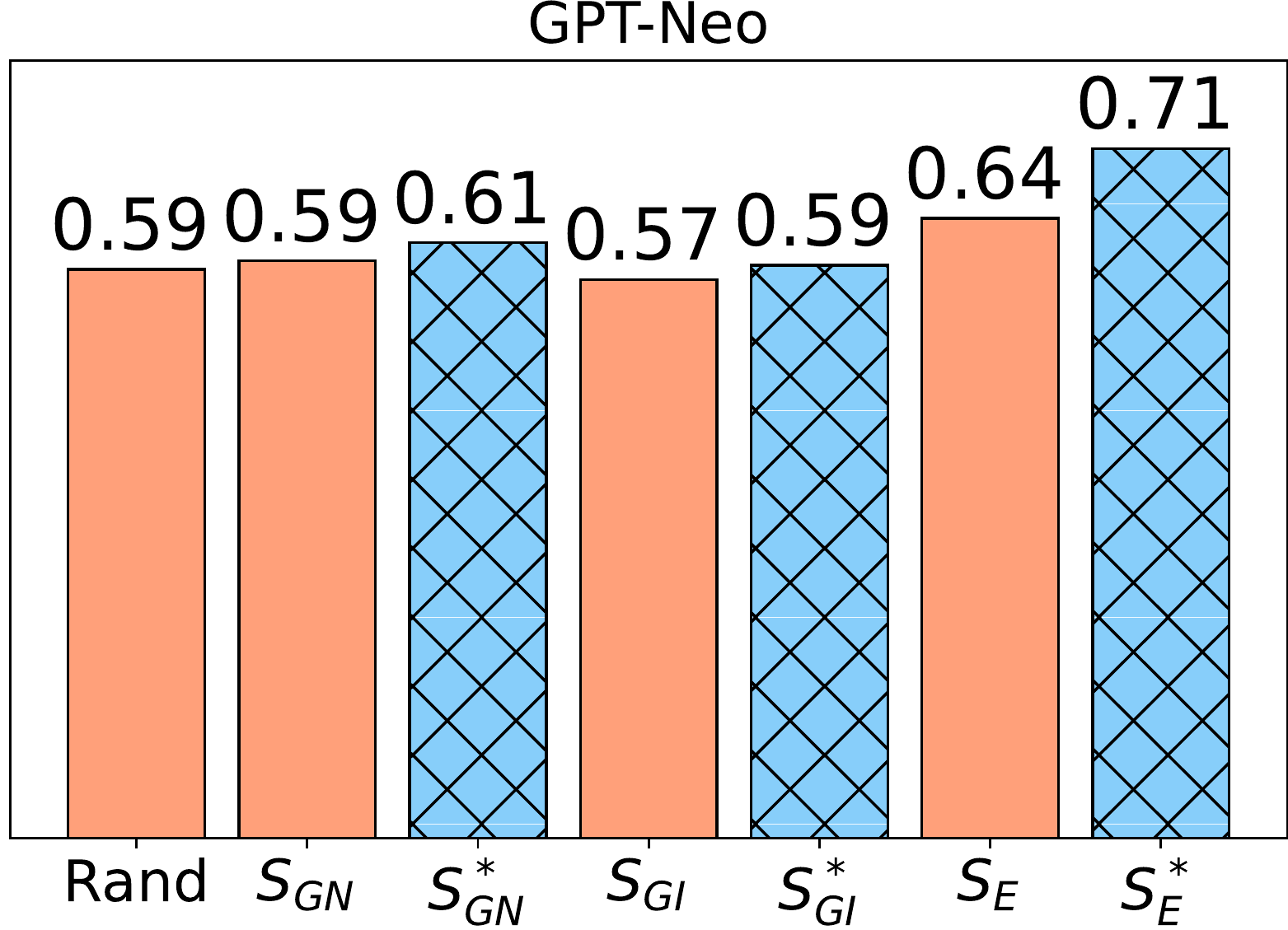}%
\subcaption{Mean Reciprocal Rank (↑)}
\end{minipage}
\caption{Alignment of different GPT-2 (left) and GPT-Neo (right) explanations with known evidence in BLiMP according to dot product (top), probes needed (middle), mean reciprocal rank (bottom) averaged over linguistic paradigms. 
}
\label{fig:blimp}
\end{figure}

\begin{figure}[htp!]
\centering
\begin{subfigure}{.5\linewidth}
  \centering
  \includegraphics[width=\linewidth]{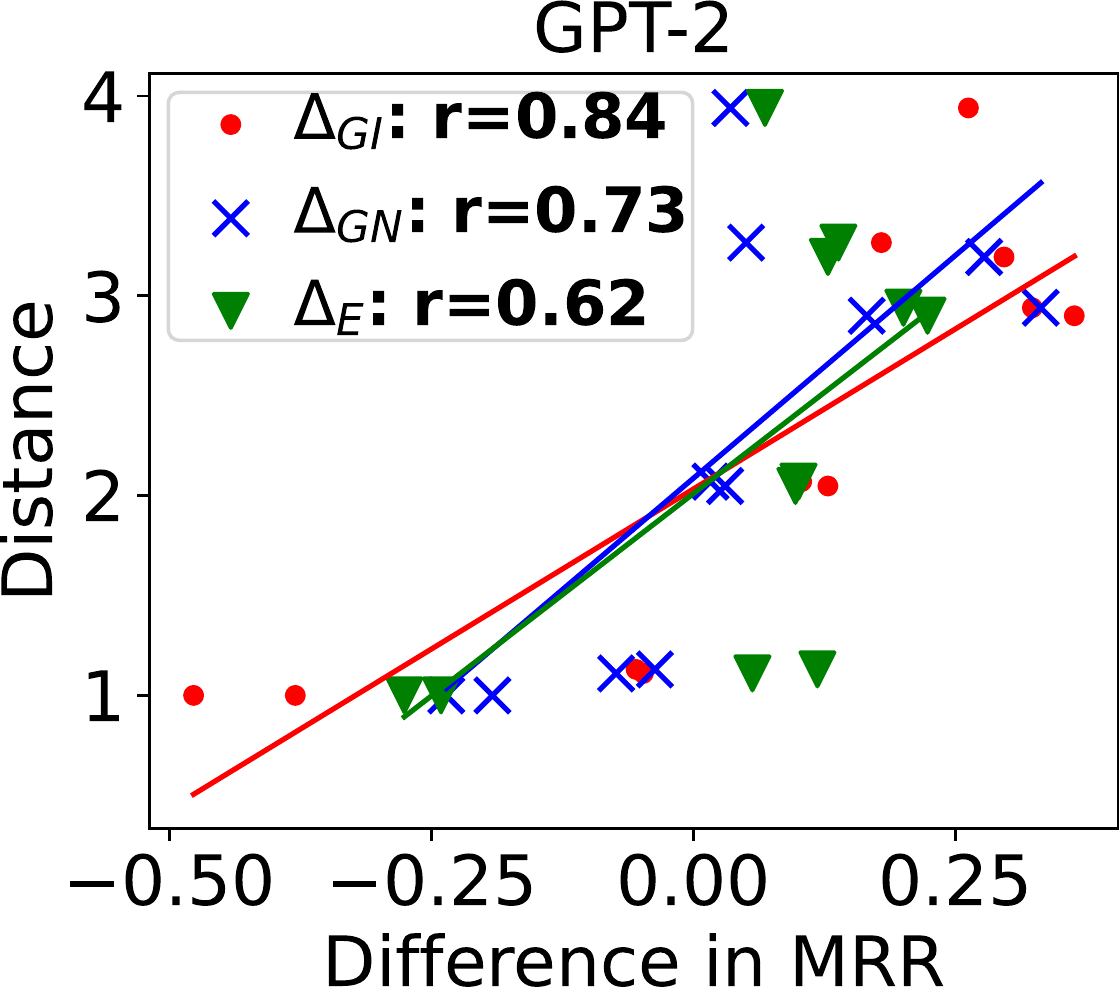}
\end{subfigure}%
\begin{subfigure}{.5\linewidth}
  \centering
  \includegraphics[width=\linewidth]{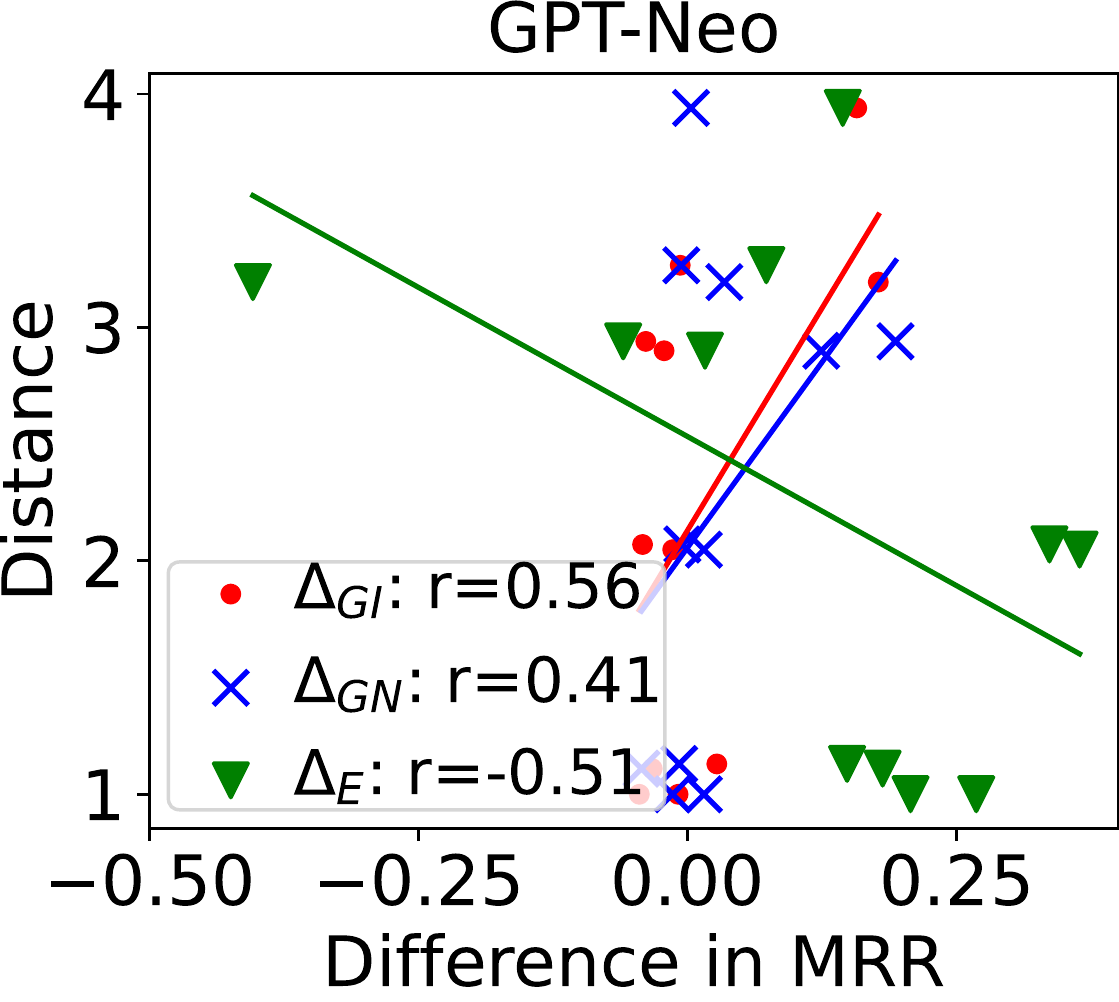}
\end{subfigure}
\caption{Scatter plot of the average distance of the known evidence to the target token across each linguistic paradigm against the difference in MRR scores between the contrastive and non-contrastive versions of each explanation method, with the Pearson correlation for each explanation method. Statistically significant Pearson's r values (p $< 0.05$) are in \textbf{bold}. In most cases, there is a positive correlation between the increase in MRR and the distance of the evidence.}
\label{fig:corr}
\end{figure}

\subsection{Results}

We use GPT-2 \cite{radford2019language} and GPT-Neo \cite{gpt-neo} to extract LM explanations. GPT-2 is a large autoregressive transformer-based LM with 1.5 billion parameters and trained on 8 million web pages. GPT-Neo is a similar LM with 2.7 billion parameters and trained on The Pile \cite{pile} which contains 825.18GB of largely English text.
In addition to the explanation methods described above, we also set up a random baseline as a comparison, where we create a vector of the same size as the explanations and the values are random samples from a uniform distribution over $[0,1)$.

In Figure \ref{fig:blimp}, we can see that overall, contrastive explanations have a higher alignment with linguistic paradigms than their non-contrastive counterparts for both GPT-2 and GPT-Neo across the different metrics.
Although non-contrastive explanations do not always outperform the random baseline, contrastive explanations have a better alignment with BLiMP than random vectors for most cases. 

In Figure \ref{fig:corr}, we see that for most explanation methods, the larger the distance between the known evidence and the target token, the larger the increase in alignment of contrastive explanations over non-contrastive explanations.  This suggests that contrastive explanations particularly outperform non-contrastive ones when the known evidence is relatively further away from the target token, that is, contrastive explanations can better capture model decisions requiring longer-range context.

In Appendix \ref{sec:appendix}, we also provide a table with the full alignment scores for each paradigm, explanation method, metric and model. 

\section{Do Contrastive Explanations Help Users Predict LM Behavior?}
\label{sec:human}

To further evaluate the quality of different explanation methods, we next describe methodology and experiments to measure to what extent explanations can improve the ability of users to predict the output of the model given its input and explanation, namely model \textit{simulatability} \cite{lipton, doshi2017towards}.

\subsection{Study Setup}
\begin{figure}[ht]
\centering
  \includegraphics[width=\linewidth]{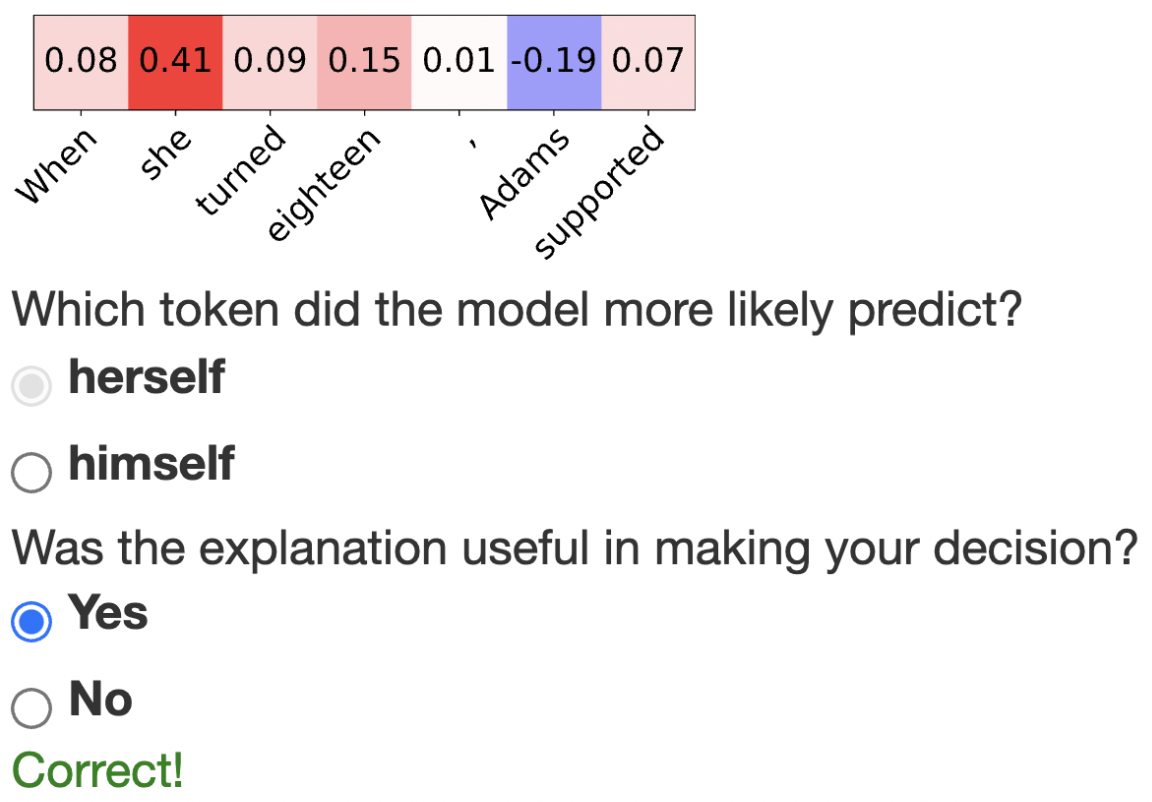}%
\caption{Example of a prompt in our human study. }
\label{fig:annot}
\end{figure}

Our user study is similar in principle to previous works that measure model simulatability given different explanations \cite{chandrasekaran-etal-2018-explanations, hase-bansal-2020-evaluating, pruthi2020evaluating}. 
In our study (Figure \ref{fig:annot}), users are given the input sequence of a GPT-2 model, two choices for the next token of the input, and the explanation for the model output. They are asked to select which of the two choices is more likely the model output, then answer whether the explanation was useful in making their decision\footnote{Although \citet{hase-bansal-2020-evaluating} have suggested not showing explanations for certain methods at test time due to potential for directly revealing the model output, this is less of a concern for saliency-based methods as their design makes it non-trivial to leak information in this way. We opt to show explanations to measure whether they sufficiently help the user make a prediction similar to the model on an individual example.}. 


We compared the effect of having no given explanation, explanations with Gradient $\times$ Input, Contrastive Gradient $\times$ Input, Erasure and Contrastive Erasure. We did not include Gradient Norm and Contrastive Gradient Norm because (1) these methods do not provide information on directionality; (2) according to Figure \ref{fig:blimp}, the alignment of known evidence in BLiMP with (Contrastive) Gradient Norm is not as good as with other explanation methods according to the number of probes needed and MRR. For non-contrastive methods, we provide the explanation for why the model predicted a token. For contrastive methods, we provide the explanation for why the model predicted one token instead of another.

We included 20 pairs of highly confusable words for users to distinguish from in our study (Appendix \ref{sec:pairs}). 10 of these word pairs were selected from the BLiMP dataset to reflect a certain linguistic phenomenon, and the other 10 word pairs were selected from pairs with the highest ``confusion score'' on the WikiText-103 test split \cite{merity2016pointer}. We define the confusion using the joint probability of a confusion from token $a$ to token $b$ given a corpus $X$:
\begin{align*}
    P(x_{true}&=a, x_{model}=b) = \\
   & \frac{1}{N} \sum_{x \in X}\sum_{t \in \text{pos}(x) | x_t = a} P_{model}(\hat{x}_t = b | x_{<t})
\end{align*}
where $x$ is a sentence in $X$, $\text{pos}(x)$ is the set of positions of tokens in $x$, $N$ is the size of the corpus. The confusion from $a$ to $b$ is the sum of the probabilities assigned by the model to token $b$ where token $a$ is the ground truth, normalized by the number of sentences in the corpus. 

The confusion score for word pair $(a,b)$ is then defined as the minimum of confusion from $a$ to $b$ and vice-versa, to ensure that both words are mutually confusable:
\begin{align*}
    \mathcal{C}(a, b) = \min(&P(x_{true}=a, x_{model}=b), \\
    &P(x_{true}=b, x_{model}=a)). \\
\end{align*}

We recruited 10 graduate students in machine learning (not authors of this paper) to perform the study. Each participant was given 10 different word pairs to annotate. For each word pair, the participant was given 40 sentences in a row, where one explanation method was chosen at random to generate the accompanying explanations. We balanced the data so that there were an equal number of examples where the true output $x_t=a$ and $x_t=b$, and also by model correctness so that the model chooses the correct output 50\% of the time, preventing users from guessing model behavior by selecting a certain token or the true token. In total, we obtained 4000 data points for model simulatability. 

\subsection{Results}

In Table \ref{table:user_accuracy}, we provide the results of our user study. For each explanation method evaluated, we computed the simulation accuracy over all samples (Acc.) as well as accuracy over samples where the model output is equal to the ground truth (Acc. Correct) and different from the ground truth (Acc. Incorrect). We also computed the percentage of explanations that users reported useful, as well as the simulation accuracy over samples where the user found the given explanation useful (Acc. Useful) and not useful (Acc. Not Useful). 

To test our results for statistical significance and account for variance in annotator skill and word pair difficulty, we fitted linear mixed-effects models using Statsmodels \cite{seabold2010statsmodels} with the annotator and word pair as random effects, the explanation method as fixed effect, and the answer accuracy or usefulness as the dependent variable. In Appendix \ref{sec:statsmodels} we provide the results of the mixed-effects models we fitted.

\paragraph{Accuracy}
\begin{table}[htp]
\centering
\resizebox{\linewidth}{!}{ 
\begin{tabular}{c|ccc|ccc}
\toprule
& & Acc.  & Acc.  &   & Acc.  & Acc.  \\
& Acc. & Correct & Incorrect &  Useful & Useful & Not Useful \\
\midrule
None & 61.38 & 74.50 & 48.25 & \textendash & \textendash & \textendash \\
$S_{GI}$ & 64.00 & 78.25 & 49.75 & 62.12 & 67.20 & 58.75 \\
$S^*_{GI}$ & \textbf{65.62} & \textbf{79.00} & 52.25 & \textbf{63.88} & 69.67 & 58.48 \\
$S_{E}$ & 63.12 & \textbf{79.00} & 47.25 & 46.50 & 65.86 & 60.75 \\
$S^*_{E}$ & \textbf{64.62} & 77.00 & 52.25 & \textbf{64.88} & 70.52 & 53.74 \\
\bottomrule
\end{tabular}}
\caption{Simulation accuracy (\%) in predicting GPT-2 outputs and subjective usefulness of explanations for various explanation methods. For each explanation method, scores that are statistically significantly higher (p $\leq$ 0.05) than the analogous method with a different contrastive setting are bolded. Overall, users achieve higher simulation accuracy with contrastive explanations.}
\label{table:user_accuracy}
\end{table}

First of all, users have the lowest accuracy in predicting the LM's output when no explanation is given, which suggests that all four types of explanations help users simulate model behavior. For both explanation methods, the contrastive setting leads to a significantly higher contrastive simulation accuracy than the non-contrastive setting.

\paragraph{Usefulness}
Contrastive explanations were also considered useful to users for model simulation significantly more often than non-contrastive explanations, with a particularly large gain in the erasure-based setting. Answer accuracy on samples where the users found the explanation useful is higher than the accuracy over all samples for each explanation method, which suggests that users can also identify useful explanations to some extent. 

These results, on the whole, provide evidence that contrastive explanations help human observers simulate model predictions more accurately.

\begin{table*}[ht!]
\centering
\resizebox{\linewidth}{!}{ 
\begin{tabular}{p{2cm}|c|p{6cm}|p{6cm}|p{6cm}}
\toprule
Phenomenon / POS & Target & Foil Cluster & Embd Nearest Neighbors & Example\\
\midrule
Anaphor Agreement & he & \textul{she}, her, She, Her, herself, hers &
 \textbf{she},She, \textbf{her}, \textbf{She}, he, they, \textbf{Her}, we, it,she, I, that,Her, you, was, there,He, is, as, in'
& That night \hlc[red!15]{,} Ilsa confronts Rick in the deserted café . When \hlc[red!50]{he} refuses to give her the letters , \_\_\_\_\_
\\
\midrule
Animate \qquad Subject & man & \textul{fruit}, mouse, ship, acid, glass, water, tree, honey, sea, ice, smoke, wood, rock, sugar, sand, cherry, dirt, fish, wind, snow
& \textbf{fruit}, fruits, Fruit, meat, flower,fruit, tomato, vegetables, \textbf{fish}, apple, berries, food, citrus, banana, vegetable, strawberry, fru, delicious, juice, foods
& You may not be surprised to learn that \hlc[red!15]{Kelly} Pool was neither \hlc[red!50]{invented} by a \_\_\_\_\_ \\
\midrule
Determiner-Noun \qquad Agreement & page & tabs, \textul{pages}, icons, stops, boxes, doors, shortcuts, bags, flavours, locks, teeth, ears, tastes, permissions, stairs, tickets, touches, cages, saves, suburbs
&  \textbf{tabs}, tab, Tab, apps, files, bags, tags, websites, sections, browsers, browser, icons, buttons, \textbf{pages}, keeps, clips, updates, 28, insists, 14
& Immediately \hlc[red!15]{after} "Heavy Competition" first aired, NBC created \hlc[red!50]{a} sub- \_\_\_\_\_
\\
\midrule
Subject-Verb Agreement & go & doesn, \textul{causes}, looks, needs, makes, isn, says, seems, seeks, displays, gives, wants, takes, uses, fav, contains, keeps, sees, tries, sounds
&  \textbf{doesn}, \textbf{isn}, didn, does, hasn, wasn, don, wouldn, \textbf{makes}, gets, has, is, aren, \textbf{gives}, Doesn, couldn, \textbf{seems}, \textbf{takes}, \textbf{keeps},doesn
& Mala \hlc[red!50]{and} the \hlc[red!15]{Eskimos} \_\_\_\_\_ \\
\midrule
ADJ & black & Black, white, black, White, \textul{red}, BLACK, green, brown, dark, orange, African, blue, yellow, pink, purple, gray, grey, whites, Brown, silver
&  \textbf{Black},Black, \textbf{black},black, \textbf{White}, \textbf{BLACK}, \textbf{white}, Blue, Red,White, In, B, The,The, It, \textbf{red}, Dark, 7, Green, \textbf{African}
& Although general \hlc[red!50]{relativity} can be used to \hlc[red!0]{perform} a \hlc[red!15]{semi} @-@ classical calculation of  \_\_\_\_\_
\\
\midrule
ADJ & black & \textul{Asian}, Chinese, English, Italian, American, Indian, East, South, British, Japanese, European, African, Eastern, North, Washington, US, West, Australian, California, London
&  \textbf{Asian},Asian, Asia, Asians, \textbf{Chinese}, \textbf{African}, \textbf{Japanese}, Korean, China, \textbf{European}, \textbf{Indian}, ethnic,Chinese, Japan, \textbf{American}, Caucasian, \textbf{Australian}, Hispanic, white, Arab
& \hlc[red!0]{While} taking part in the American \hlc[red!15]{Negro} Academy (ANA) in 1897 , Du Bois presented a paper in which he rejected Frederick \hlc[red!50]{Douglass} 's plea for \_\_\_\_\_ 
\\
\midrule
ADP & for & \textul{to}, in, and, on, with, for, when, from, at, (, if, as, after, by, over, because, while, without, before, through
&  \textbf{to}, \textbf{in}, \textbf{for}, \textbf{on}, \textbf{and}, \textbf{as}, \textbf{with}, of, a, \textbf{at}, that, the, \textbf{from}, \textbf{by}, an, (, To, is, it, or
& The \hlc[red!15]{war} of words \hlc[red!0]{would} \hlc[red!50]{continue} \_\_\_\_\_
\\
\midrule
ADV & back & the, to, a, \textul{in}, and, on, of, it, ", not, that, with, for, this, from, up, just, at, (, all
&  \textbf{the}, \textbf{a}, an, \textbf{it}, \textbf{this}, \textbf{that}, \textbf{in}, The, \textbf{to},The, \textbf{all}, \textbf{and}, their, as, \textbf{for}, \textbf{on}, his, \textbf{at}, some, what
& \hlc[red!15]{One} would have thought that \hlc[red!0]{claims} \hlc[red!50]{dating} \_\_\_\_\_
\\
\midrule
DET & his & \textul{the}, you, it, not, that, my, [, this, your, he, all, so, what, there, her, some, his, time, him, He
&  \textbf{the}, a, an, \textbf{it}, \textbf{this}, \textbf{that}, in, The, to,The, \textbf{all}, and, their, as, for, on, \textbf{his}, at, \textbf{some}, \textbf{what}
& A preview screening of Sweet Smell of Success was poorly received , as \hlc[red!15]{Tony} \hlc[red!50]{Curtis} fans were expecting him to \hlc[red!0]{play} one of \_\_\_\_\_
\\
\midrule
NOUN & girl & Guy, \textul{Jack}, Jones, Robin, James, David, Tom, Todd, Frank, Mike, Jimmy, Michael, Peter, George, William, Bill, Smith, Tony, Harry, Jackson
&   \textbf{Guy},Guy, guy,guy, Gu, Dave, Man, dude, Girl, Guys, John, Steve, \textbackslash x00, \textbackslash xef \textbackslash xbf \textbackslash xbd, \textbackslash xef \textbackslash xbf \textbackslash xbd, \textbackslash x1b, \textbackslash xef \textbackslash xbf \textbackslash xbd, \textbackslash x12, \textbackslash x1c, \textbackslash x16
& \hlc[red!50]{Veronica} \hlc[red!0]{talks} to to Sean Friedrich and \hlc[red!15]{tells} him about the \_\_\_\_\_
\\
\midrule
NUM & five & \textul{the}, to, a, in, and, on, of, is, it, ", not, that, 1, with, for, 2, this, up, just, at
&  \textbf{the}, \textbf{a}, an, \textbf{it}, \textbf{this}, \textbf{that}, \textbf{in}, The, \textbf{to},The, all, \textbf{and}, their, as, \textbf{for}, \textbf{on}, his, \textbf{at}, some, what
& \hlc[red!15]{From} \hlc[red!0]{the} \hlc[red!50]{age} of \_\_\_\_\_
\\
\midrule
VERB & going & \textul{got}, didn, won, opened, told, went, heard, saw, wanted, lost, came, started, took, gave, happened, tried, couldn, died, turned, looked
&  \textbf{got}, gets, get, had, \textbf{went}, \textbf{gave}, \textbf{took}, \textbf{came}, \textbf{didn}, did, getting, been, became, has, was, made, \textbf{started}, have, gotten, showed
&\hlc[red!0]{Truman} \hlc[red!15]{had} dreamed \hlc[red!50]{of} \_\_\_\_\_
\\
\bottomrule
\end{tabular}}
\caption{Examples of foil clusters obtained by clustering contrastive explanations of GPT-2. For each cluster, the 20 most frequent foils are shown, as well as the 20 nearest neighbors in the word embedding space of the first foil, and an example is included for the contrastive explanation of the target token vs. the underlined foil in the cluster. In each explanation, the two most salient input tokens are highlighted in decreasing intensity of red. 
}
\label{table:clusters}
\end{table*}

\section{What Context Do Models Use for Certain Decisions?}
\label{sec:cluster}


Finally, we use contrastive explanations to discover how language models achieve various linguistic distinctions. We hypothesize that \emph{similar evidence is necessary to disambiguate foils that are similar linguistically}. To test this hypothesis, we propose a methodology where we first represent each token by a vector representing its saliency map when the token is used as a foil in contrastive explanation of a particular target word. Conceptually, this vector represents the \emph{type of context} that is necessary to disambiguate the particular token from the target. Next, we use a clustering algorithm on these vectors, generating clusters of foils where similar types of context are useful to disambiguate. We then verify whether we find clusters associated with \emph{salient linguistic distinctions} defined a-priori. Finally, we inspect the mean vectors of explanations associated with foils in the cluster to investigate how models perform these linguistic distinctions. 

\subsection{Methodology}

We generate contrastive explanations for the 10 most frequent words in each major part of speech as the target token, and use the 10,000 most frequent vocabulary items as foils. For each target $y_t$, we randomly select 500 sentences from the WikiText-103 dataset \cite{merity2016pointer} and obtain a sentence set $X$. Then, for each foil $y_f$ and each sentence $x_i \in X$, we generate a single contrastive explanation $e(x_i, y_t, y_f)$. Then, for each target $y_t$ and foil $y_f$, we generate an aggregate explanation vector $e(y_t, y_f) = \bigoplus_{x_i \in X} e(x_i, y_t, y_f)$ by concatenating the single explanation vectors for each sentence in the corpus.

Then, for a given target $y_t$, we apply k-means clustering on the concatenated contrastive explanations associated with different foils $y_f$ to cluster foils by explanation similarity. We use GPT-2 to extract all the contrastive explanations due to its better alignment with linguistic phenomena than GPT-Neo (\S\ref{sec:blimp-eval}). We only extract contrastive explanations with gradient norm and gradient$\times$input due to the computational complexity of input erasure (\S\ref{sec:erasure}). 

In Table \ref{table:clusters}, we show examples of the obtained foil clusters. The foils in each cluster are in descending frequency in training data. For the most frequent foil in each cluster, we also retrieve its 20 nearest neighbors in the word embedding space according to the Minkowski distance to compare them with clusters of foils. 



\subsection{Foil Clusters}
First, we find that linguistically similar foils are indeed clustered together: we discover clusters relating to a variety of previously studied linguistic phenomena, a few of which we detail below and give examples in Table \ref{table:clusters}. Moreover, foil clusters reflect linguistic distinctions that are not found in the nearest neighbors of word embeddings. This suggests that the model use similar types of input features to make certain decisions.


\textbf{Anaphor agreement:} To predict anaphor agreement, models must contrast pronouns from other pronouns with different gender or number. We find that indeed, when the target is a pronoun, other pronouns of a different gender or number are often clustered together: when the target is a male pronoun, we find a cluster of female pronouns. The foil cluster containing ``she'' includes several types of pronouns that are all of the female gender. On the other hand, the nearest neighbors of ``she'' are mostly limited to subject and object pronouns, and they are of various genders and numbers.

\textbf{Animacy:} In certain verb phrases, the main verb enforces that the subject is animate. Reflecting this, when the target is an animate noun, inanimate nouns form a cluster. While the foil cluster in Table \ref{table:clusters} contains a variety of singular inanimate nouns, the nearest neighbors of ``fruit'' are mostly both singular and plural nouns related to produce. 

\textbf{Plurality:} For determiner-noun agreement, singular nouns are contrasted with clusters of plural noun foils, and vice-versa. We find examples of clusters of plural nouns when the target is a singular noun, whereas the nearest neighbors of ``tabs'' are both singular and plural nouns. To verify subject-verb agreement, when the target is a plural verb, singular verbs are clustered together, but the nearest neighbors of ``doesn'' contain both singular and plural verbs, especially negative contractions. 


\subsection{Explanation Analysis Results}

By analyzing the explanations associated with different clusters, we are also able to learn various interesting properties of how GPT-2 makes certain predictions. We provide the full results of our analysis in Appendix \ref{sec:analysis}. 

For example, to distinguish between adjectives, the model often relies on input words that are semantically similar to the target: to distinguish ``black'' from other colors, words such as ``relativity'' are important. To contrast adpositions and adverbs from other words with the same POS, verbs in the input that are associated with the target word are useful: for example, the verbs ``dating'' and ``traced'' are useful when the target is ``back''. 

To choose the correct gender for determiners, nouns and pronouns, the model often uses common proper nouns such as ``Veronica'' and other gendered words such as ``he'' in the input. To disambiguate numbers from non-number words, input
words related to enumeration or measurement (e.g. ``age'', ``consists'', ``least'') are useful. When the target word is a verb, other verbs in other verb forms are often clustered together, which suggests that the model uses similar input features to verify subject-verb agreement.

Our analysis also reveals why the model may have made certain mistakes. For example, when the model generates a pronoun of the incorrect gender, often the model was influenced by a gender neutral proper noun in the input, or by proper nouns and pronouns of the opposite gender that appear in the input.

Overall, our methodology for clustering contrastive explanations provides an aggregate analysis of linguistic distinctions to understand general properties of language model decisions.



\section{Conclusion and Future Work}
In this work, we interpreted language model decisions using contrastive explanations by extending three existing input saliency methods to the contrastive setting. We also proposed three new methods to evaluate and explore the quality of contrastive explanations: an alignment evaluation to verify whether explanations capture linguistically appropriate evidence, a user evaluation to measure the contrastive model simulatability of explanations, and a clustering-based aggregate analysis to investigate model properties using contrastive explanations. 

We find that contrastive explanations are better aligned to known evidence related to major grammatical phenomena than their non-contrastive counterparts. Moreover, contrastive explanations allow better contrastive simulatability of models for users. From there, we studied what kinds of decisions require similar evidence and we used contrastive explanations to characterize how models make certain linguistic distinctions. Overall, contrastive explanations give a more intuitive and fine-grained interpretation of language models. 

Future work could explore the application of these contrastive explanations to other machine learning models and tasks, extending other interpretability methods to the contrastive setting, as well as using what we learn about models through contrastive explanations to improve them.

\section*{Acknowledgements}
We would like to thank Danish Pruthi and Paul Michel for initial discussions about this paper topic, Lucio Dery and Patrick Fernandes for helpful feedback. We would also like to thank Cathy Jiao, Clara Na, Chen Cui, Emmy Liu, Karthik Ganesan, Kunal Dhawan, Shubham Phal, Sireesh Gururaja and Sumanth Subramanya Rao for participating in the human evaluation. This work was supported by the Siebel Scholars Award and the CMU-Portugal MAIA program.

\bibliography{anthology,custom}
\bibliographystyle{acl_natbib}

\appendix

\section{Contrastive Explanations for Neural Machine Translation (NMT) Models}
\label{sec:nmt}

\subsection{Extending Contrastive Explanations to NMT}

Machine translation can be thought of as a specific type of language models where the model is conditioned on both the source sentence and the partial translation. It has similar complexities as monolingual language modeling that make interpreting neural machine translation (NMT) models difficult. We therefore also extend contrastive explanations to NMT models. 

We compute the contrastive gradient norm saliency for an NMT model by first calculating the gradient over the encoder input (the source sentence) and over the decoder input (the partial translation) as:
$$
g^*(x_i^e) = \nabla_{x_i^e} \left( q(y_t | \bm{x^e}, \bm{x^d}) - q(y_f | \bm{x^e}, \bm{x^d}) \right) $$$$
g^*(x_i^d) = \nabla_{x_i^d} \left( q(y_t | \bm{x^e}, \bm{x^d}) - q(y_f | \bm{x^e}, \bm{x^d}) \right)
$$ 

where $\bm{x^e}$ is the encoder input, $\bm{x^d}$ is the decoder input, and the other notations follow the ones in \S\ref{sec:norm}.

Then, the contrastive gradient norm for each $x_i^e$ and $x_i^d$ are:
$$
S_{GN}^*(x_i^e) = ||g^*(x_i^e)||_{L1} $$$$
S_{GN}^*(x_i^d) = ||g^*(x_i^d)||_{L1}
$$

Similarly, the contrastive gradient $\times$ input are:
$$
S_{GI}^*(x_i^e) = g^*(x_i^e) \cdot x_i^e $$$$
S_{GI}^*(x_i^d) = g^*(x_i^d) \cdot x_i^d
$$

We define the input erasure for each $x_i^e$ and $x_i^d$ as:
\begin{align*}
    S^*_{E}(x_i^e) = &\left(q(y_t | \bm{x^e}, \bm{x^d}) - q(y_t | \bm{x^e}_{\neg i}, \bm{x^d})\right) \\
    &- \left(q(y_f | \bm{x^e}, \bm{x^d}) - q(y_f | \bm{x^e}_{\neg i}, \bm{x^d})\right) 
\end{align*}
\begin{align*}
S^*_{E}(x_i^d) = &\left(q(y_t | \bm{x^e}, \bm{x^d}) - q(y_t | \bm{x^e}, \bm{x^d}_{\neg i})\right) \\
- &\left(q(y_f | \bm{x^e}, \bm{x^d}) - q(y_f | \bm{x^e}, \bm{x^d}_{\neg i})\right)
\end{align*}

\begin{table}[htp]
\centering
\resizebox{\linewidth}{!}{ 
\begin{tabular}{l} \toprule
\textbf{Why did the model predict "il" ?} \\
\textbf{en:} \hlc[blue!18]{I} \hlc[blue!7]{ordered} \hlc[red!11]{a} \hlc[red!16]{new} \hlc[red!59]{vase} \hlc[red!35]{and} \hlc[blue!18]{it} \hlc[blue!16]{arrived} \hlc[red!17]{today}  \\
\textbf{fr:} \hlc[blue!14]{J} \hlc[red!4]{'} \hlc[blue!13]{ai} \hlc[red!14]{commandé} \hlc[blue!8]{un} \hlc[blue!129]{nouveau} \hlc[blue!40]{vase} \hlc[blue!42]{et}  \\
\textbf{Why did the model predict "il" instead of "elle" ?} \\
\textbf{en:} \hlc[blue!6]{I} \hlc[blue!45]{ordered} \hlc[blue!21]{a} \hlc[blue!8]{new} \hlc[red!43]{vase} \hlc[blue!29]{and} \hlc[red!12]{it} \hlc[blue!26]{arrived} \hlc[blue!10]{today}  \\
\textbf{fr:} \hlc[blue!0]{J} \hlc[red!4]{'} \hlc[red!8]{ai} \hlc[red!17]{commandé} \hlc[blue!4]{un} \hlc[red!28]{nouveau} \hlc[red!26]{vase} \hlc[blue!59]{et}  \\
\midrule
\textbf{2. Why did the model predict "votre" ?} \\
\textbf{en:} \hlc[red!6]{You} \hlc[red!71]{cannot} \hlc[red!10]{bring} \hlc[red!51]{your} \hlc[red!46]{dog} \hlc[blue!21]{here} \hlc[red!15]{.}  \\
\textbf{fr:} \hlc[blue!12]{Vous} \hlc[blue!4]{ne} \hlc[blue!78]{pouvez} \hlc[blue!21]{pas} \hlc[blue!35]{amener}  \\
\textbf{Why did the model predict "votre" instead of "ton" ?} \\
\textbf{en:} \hlc[blue!12]{You} \hlc[red!30]{cannot} \hlc[red!22]{bring} \hlc[red!88]{your} \hlc[red!60]{dog} \hlc[blue!21]{here} \hlc[red!9]{.}  \\
\textbf{fr:} \hlc[red!38]{Vous} \hlc[red!0]{ne} \hlc[red!41]{pouvez} \hlc[blue!26]{pas} \hlc[red!21]{amener}  \\
\midrule
\textbf{3. Why did the model predict "apprenais" ?} \\
\textbf{en:} \hlc[red!5]{I} \hlc[blue!8]{liked} \hlc[red!10]{school} \hlc[red!19]{because} \hlc[red!32]{I} \hlc[blue!47]{learned} \hlc[red!30]{a} \hlc[blue!8]{lot} \hlc[red!5]{there} \hlc[blue!2]{.}  \\
\textbf{fr:} \hlc[red!4]{J} \hlc[blue!20]{'} \hlc[blue!37]{aimais} \hlc[blue!3]{l} \hlc[blue!6]{'} \hlc[blue!8]{école} \hlc[blue!18]{parce} \hlc[blue!46]{que} \hlc[blue!31]{j} \hlc[blue!47]{'}  \\
\textbf{Why did the model predict "apprenais" instead of "ai" ?} \\
\textbf{en:} \hlc[red!13]{I} \hlc[blue!15]{liked} \hlc[red!6]{school} \hlc[blue!7]{because} \hlc[red!29]{I} \hlc[blue!6]{learned} \hlc[blue!17]{a} \hlc[red!12]{lot} \hlc[blue!30]{there} \hlc[blue!20]{.}  \\
\textbf{fr:} \hlc[blue!8]{J} \hlc[red!2]{'} \hlc[red!7]{aimais} \hlc[blue!9]{l} \hlc[red!4]{'} \hlc[red!10]{école} \hlc[blue!17]{parce} \hlc[red!4]{que} \hlc[blue!126]{j} \hlc[red!59]{'}  \\
\midrule
\textbf{4. Why did the model predict "sais" ?} \\
\textbf{en:} \hlc[red!3]{They} \hlc[red!14]{know} \hlc[red!7]{what} \hlc[blue!3]{to} \hlc[red!1]{do} \hlc[blue!9]{,} \hlc[red!12]{I} \hlc[blue!15]{don} \hlc[blue!15]{'} \hlc[red!17]{t} \hlc[blue!20]{.}  \\
\textbf{fr:} \hlc[blue!8]{Ils} \hlc[blue!84]{savent} \hlc[red!126]{quoi} \hlc[blue!48]{faire} \hlc[blue!18]{,} \hlc[blue!7]{je} \hlc[red!6]{ne}  \\
\textbf{Why did the model predict "sais" instead of "veux" ?} \\
\textbf{en:} \hlc[red!5]{They} \hlc[red!3]{know} \hlc[red!15]{what} \hlc[blue!0]{to} \hlc[red!2]{do} \hlc[blue!1]{,} \hlc[red!27]{I} \hlc[red!4]{don} \hlc[blue!9]{'} \hlc[red!16]{t} \hlc[blue!5]{.}  \\
\textbf{fr:} \hlc[blue!9]{Ils} \hlc[red!81]{savent} \hlc[blue!92]{quoi} \hlc[blue!18]{faire} \hlc[red!5]{,} \hlc[red!23]{je} \hlc[blue!51]{ne}  \\
\midrule
\textbf{5. Why did the model predict "carnet" ?} \\
\textbf{en:} \hlc[blue!2]{I} \hlc[red!5]{like} \hlc[red!1]{my} \hlc[red!4]{old} \hlc[red!34]{notebook} \hlc[blue!2]{better} \hlc[red!7]{than} \hlc[red!1]{my} \hlc[red!3]{new} \hlc[red!152]{notebook}  \\
\textbf{fr:} \hlc[red!2]{J} \hlc[blue!3]{'} \hlc[red!1]{aime} \hlc[blue!19]{mieux} \hlc[red!1]{mon} \hlc[red!15]{ancien} \hlc[blue!5]{carnet} \hlc[blue!1]{que} \hlc[blue!12]{mon} \hlc[red!78]{nouveau}  \\
\textbf{Why did the model predict "carnet" instead of "ordinateur" ?} \\
\textbf{en:} \hlc[blue!12]{I} \hlc[blue!8]{like} \hlc[blue!6]{my} \hlc[blue!21]{old} \hlc[red!51]{notebook} \hlc[blue!17]{better} \hlc[red!0]{than} \hlc[blue!10]{my} \hlc[red!11]{new} \hlc[blue!16]{notebook}  \\
\textbf{fr:} \hlc[red!16]{J} \hlc[red!7]{'} \hlc[blue!14]{aime} \hlc[blue!58]{mieux} \hlc[red!2]{mon} \hlc[blue!1]{ancien} \hlc[red!36]{carnet} \hlc[red!19]{que} \hlc[red!5]{mon} \hlc[blue!13]{nouveau}  \\
\bottomrule
\end{tabular}}
\caption{Examples of non-contrastive and contrastive explanations for NMT models translating from English to French using input $\times$ gradient. Input tokens that are measured to raise or lower the probability of each decision are in red and blue respectively, and those with little influence are in white.}
\label{table:nmt}
\end{table}

\subsection{Qualitative Results}

In Table \ref{table:nmt}, we provide examples of non-contrastive and contrastive explanations for NMT decisions. We use MarianMT \cite{mariannmt} with pre-trained weights from the model trained to translate from English to Romance languages\footnote{\url{https://github.com/Helsinki-NLP/Tatoeba-Challenge/blob/master/models/eng-roa/README.md}} to extract explanations. Each example reflects a decision associated with one of the five types of linguistic ambiguities during translation identified in \citet{yin2021muda}. 

In the first example, the model must translate the gender neutral English pronoun ``it'' into the masculine French pronoun ``il''. In both non-contrastive and contrastive explanations, the English antecedent ``vase'' influences the model to predict ``il'', however to disambiguate ``il'' from the feminine pronoun ``elle'', the model also relies on the french antecedent and its masculine adjective ``nouveau vase''. 

In the second example, the model must translate ``your'' with the formality level consistent with the partial translation. While in the non-contrastive explanation, only tokens in the source sentence are salient which do not explain the model's choice of formality level, in the contrastive explanation, other French words in the polite formality level such as ``Vous'' and ``pouvez'' are salient.

In the third example, the model must translate ``learned'' using the verb form that is consistent with the partial translation. Similarly to the previous example, only the contrastive explanation contains salient tokens in the same verb from as the target token such as ``aimais''.

In the fourth example, the model needs to resolve the elided verb in ``I don't \st{know}'' to translate into French. The contrastive explanation with a different verb as a foil shows that the elided verb in the target side makes the correct verb more likely than another verb.

In the fifth example, the model must choose the translation that is lexically cohesive with the partial translation, where ``carnet'' refers to a book with paper pages and ``ordinateur'' refers to a computer notebook. In the non-contrastive explanation, the word ``notebook'' and the target token preceding the prediction are the most salient. In the contrastive explanation, the word ``carnet'' in the partial translation also becomes salient.

\section{Alignment of Contrastive Explanations to Linguistic Paradigms}
\label{sec:appendix}
In Table \ref{table:align}, we present the full alignment scores of contrastive explanations from GPT-2 and GPT-Neo models with the known evidence to disambiguate linguistic paradigms in the BLiMP dataset.

\begin{table*}[htp]
\centering
\resizebox{\linewidth}{!}{ 
\begin{tabular}{ccc|ccc|ccc}
\toprule
& & & \multicolumn{3}{c}{GPT-2} & \multicolumn{3}{c}{GPT-Neo}\\
Paradigm & Dist & Explanation & Dot Product (↑)& Probes Needed (↓) & MRR (↑) & Dot Product (↑)& Probes Needed (↓) & MRR (↑) \\
\midrule
\multirow{7}{*}{anaphor\_gender\_agreement} & & Random & 0.528 & 0.706 & 0.718 & 0.548 & 0.618 & 0.762\\
& & $S_{GN}$ & 0.429 & 1.384 & 0.478 & 0.480 & 0.828 & 0.622\\
& & $S_{GN}^*$ & \textbf{0.834} & \textbf{0.472} & \textbf{0.809} & \textbf{0.785} & \textbf{0.432} & \textbf{0.815}\\
&2.94 & $S_{GI}$ &  \textbf{0.078} & 1.402 & 0.468 & \textbf{-0.054} & \textbf{0.526} & \textbf{0.786}\\
& & $S_{GI}^*$ & -0.019 & \textbf{0.502} & \textbf{0.791} & -0.133 & 0.684 & 0.747\\
& & $S_{E}$ & -0.350 & 0.564 & 0.764 & \textbf{0.645} & \textbf{0.078} & \textbf{0.963}\\
& & $S_{E}^*$ & \textbf{0.603} & \textbf{0.090} & \textbf{0.964} & 0.637 & 0.156 & 0.903\\
\midrule
\multirow{7}{*}{anaphor\_number\_agreement} & & Random & 0.554 & 0.666 & 0.741 & 0.568 & 0.598 & 0.756\\
& & $S_{GN}$ & 0.463 & 1.268 & 0.512 & 0.508 & 0.784 & 0.639\\
& & $S_{GN}^*$ & \textbf{0.841} & \textbf{0.702} & \textbf{0.677} & \textbf{0.816} & \textbf{0.524} & \textbf{0.763}\\
&2.90 & $S_{GI}$ &  0.084 & 1.346 & 0.497 & -0.095 & \textbf{0.510} & \textbf{0.797}\\
& & $S_{GI}^*$ & 0.084 & \textbf{0.408} & \textbf{0.860} & \textbf{-0.068} & 0.636 & 0.775\\
& & $S_{E}$ & -0.349 & 0.704 & 0.728 & 0.618 & 0.128 & 0.940\\
& & $S_{E}^*$ & \textbf{0.604} & \textbf{0.136} & \textbf{0.951} & \textbf{0.666} & \textbf{0.106} & \textbf{0.956}\\
\midrule
\multirow{7}{*}{animate\_subject\_passive} & & Random & 0.155 & 2.940 & 0.378 & 0.150 & 2.976 & 0.379\\
& & $S_{GN}$ & 0.211 & 1.080 & 0.699 & 0.236 & \textbf{0.828} & \textbf{0.727}\\
& & $S_{GN}^*$ & \textbf{0.463} & \textbf{0.754} & \textbf{0.749} & \textbf{0.452} & 0.862 & 0.721\\
&3.27 & $S_{GI}$ &  0.016 & 4.004 & 0.233 & \textbf{0.020} & \textbf{2.780} & \textbf{0.416}\\
& & $S_{GI}^*$ & \textbf{0.069} & \textbf{2.782} & \textbf{0.412} & 0.016 & 2.844 & 0.409\\
& & $S_{E}$ & -0.036 & 3.214 & 0.362 & \textbf{0.168} & \textbf{2.024} & 0.444\\
& & $S_{E}^*$ & \textbf{0.125} & \textbf{2.122} & \textbf{0.500} & 0.123 & 2.120 & \textbf{0.517}\\
\midrule
\multirow{7}{*}{determiner\_noun\_agreement\_1} & & Random & 0.208 & 2.202 & 0.449 & 0.207 & 2.142 & 0.461\\
& & $S_{GN}$ & 0.239 & \textbf{1.320} & \textbf{0.598} & 0.150 & 2.954 & 0.287\\
& & $S_{GN}^*$ & \textbf{0.275} & 2.680 & 0.406 & \textbf{0.258} & \textbf{2.906} & \textbf{0.302}\\
&1.00 & $S_{GI}$ &  \textbf{0.560} & \textbf{0.038} & \textbf{0.983} & \textbf{-0.042} & \textbf{2.384} & \textbf{0.380}\\
& & $S_{GI}^*$ & 0.162 & 1.558 & 0.603 & -0.056 & 2.554 & 0.371\\
& & $S_{E}$ & 0.022 & \textbf{1.150} & \textbf{0.604} & 0.234 & 1.290 & 0.543\\
& & $S_{E}^*$ & \textbf{0.031} & 2.598 & 0.363 & \textbf{0.362} & \textbf{0.612} & \textbf{0.811}\\
\midrule
\multirow{7}{*}{determiner\_noun\_agreement\_irregular\_1} & & Random & 0.198 & 2.248 & 0.437 & 0.202 & 2.110 & 0.456\\
& & $S_{GN}$ & 0.236 & \textbf{1.228} & \textbf{0.616} & 0.160 & \textbf{2.716} & \textbf{0.324}\\
& & $S_{GN}^*$ & \textbf{0.286} & 2.578 & 0.380 & \textbf{0.266} & 2.826 & 0.310\\
&1.00 & $S_{GI}$ &  \textbf{0.559} & \textbf{0.034} & \textbf{0.984} & \textbf{-0.035} & \textbf{2.160} & \textbf{0.419}\\
& & $S_{GI}^*$ & 0.046 & 2.038 & 0.507 & -0.046 & 2.428 & 0.374\\
& & $S_{E}$ & 0.020 & \textbf{1.082} & \textbf{0.628} & 0.205 & 1.360 & 0.548\\
& & $S_{E}^*$ & \textbf{0.026} & 2.502 & 0.352 & \textbf{0.306} & \textbf{0.784} & \textbf{0.755}\\
\midrule
\multirow{7}{*}{determiner\_noun\_agreement\_with\_adjective\_1} & & Random & 0.167 & 2.672 & 0.406 & 0.168 & 2.672 & 0.405\\
& & $S_{GN}$ & 0.118 & 3.914 & 0.237 & 0.120 & 3.902 & 0.230\\
& & $S_{GN}^*$ & \textbf{0.210} & \textbf{3.532} & \textbf{0.267} & \textbf{0.228} & \textbf{3.814} & \textbf{0.245}\\
&2.05 & $S_{GI}$ &  0.118 & 2.426 & 0.354 & \textbf{-0.010} & \textbf{2.736} & \textbf{0.356}\\
& & $S_{GI}^*$ & \textbf{0.141} & \textbf{2.012} & \textbf{0.482} & -0.051 & 2.950 & 0.342\\
& & $S_{E}$ & 0.042 & 1.730 & 0.583 & 0.092 & 2.748 & 0.333\\
& & $S_{E}^*$ & \textbf{0.305} & \textbf{1.084} & \textbf{0.680} & \textbf{0.260} & \textbf{1.176} & \textbf{0.697}\\
\midrule
\multirow{7}{*}{determiner\_noun\_agreement\_with\_adj\_irregular\_1} & & Random & 0.167 & 2.620 & 0.401 & 0.158 & 2.820 & 0.392\\
& & $S_{GN}$ & 0.116 & 3.920 & 0.240 & 0.125 & \textbf{3.620} & \textbf{0.248}\\
& & $S_{GN}^*$ & \textbf{0.205} & \textbf{3.664} & \textbf{0.256} & \textbf{0.228} & 3.718 & 0.243\\
&2.07 & $S_{GI}$ &  0.106 & 2.620 & 0.345 & \textbf{-0.007} & \textbf{2.754} & \textbf{0.358}\\
& & $S_{GI}^*$ & \textbf{0.111} & \textbf{2.244} & \textbf{0.448} & -0.047 & 3.126 & 0.316\\
& & $S_{E}$ & 0.048 & 1.688 & 0.586 & 0.103 & 2.644 & 0.347\\
& & $S_{E}^*$ & \textbf{0.313} & \textbf{1.024} & \textbf{0.686} & \textbf{0.263} & \textbf{1.066} & \textbf{0.683}\\
\midrule
\multirow{7}{*}{npi\_present\_1} & & Random & 0.336 & 1.080 & 0.604 & 0.350 & 0.984 & 0.632\\
& & $S_{GN}$ & 0.294 & 1.160 & 0.510 & 0.376 & 0.454 & 0.778\\
& & $S_{GN}^*$ & \textbf{0.456} & \textbf{0.450} & \textbf{0.787} & \textbf{0.449} & \textbf{0.382} & \textbf{0.812}\\
&3.19 & $S_{GI}$ &  0.100 & 1.374 & 0.463 & -0.160 & 1.288 & 0.575\\
& & $S_{GI}^*$ & \textbf{0.144} & \textbf{0.570} & \textbf{0.759} & \textbf{0.202} & \textbf{0.766} & \textbf{0.752}\\
& & $S_{E}$ & -0.336 & 1.514 & 0.556 & \textbf{0.624} & \textbf{0.086} & \textbf{0.960}\\
& & $S_{E}^*$ & \textbf{0.160} & \textbf{0.902} & \textbf{0.684} & 0.062 & 1.204 & 0.556\\
\midrule
\multirow{7}{*}{distractor\_agreement\_relational\_noun} & & Random & 0.230 & 1.936 & 0.494 & 0.227 & 2.106 & 0.463\\
& & $S_{GN}$ & 0.266 & 1.199 & 0.584 & 0.269 & \textbf{0.965} & 0.646\\
& & $S_{GN}^*$ & \textbf{0.408} & \textbf{1.092} & \textbf{0.619} & \textbf{0.392} & 1.000 & \textbf{0.649}\\
&3.94 & $S_{GI}$ &  0.044 & 2.291 & 0.369 & -0.066 & 2.326 & 0.434\\
& & $S_{GI}^*$ & \textbf{0.223} & \textbf{1.057} & \textbf{0.631} & \textbf{0.051} & \textbf{1.383} & \textbf{0.591}\\
& & $S_{E}$ & -0.023 & 1.922 & 0.434 & 0.120 & 2.007 & 0.400\\
& & $S_{E}^*$ & \textbf{0.190} & \textbf{1.709} & \textbf{0.502} & \textbf{0.186} & \textbf{1.617} & \textbf{0.544}\\
\midrule
\multirow{7}{*}{irregular\_plural\_subject\_verb\_agreement\_1} & & Random & 0.561 & 0.539 & 0.760 & 0.545 & 0.494 & 0.769\\
& & $S_{GN}$ & 0.652 & \textbf{0.242} & \textbf{0.917} & 0.610 & \textbf{0.348} & \textbf{0.860}\\
& & $S_{GN}^*$ & \textbf{0.676} & 0.315 & 0.843 & \textbf{0.644} & 0.376 & 0.817\\
&1.11 & $S_{GI}$ &  \textbf{0.590} & \textbf{0.253} & \textbf{0.912} & \textbf{0.067} & \textbf{0.472} & \textbf{0.783}\\
& & $S_{GI}^*$ & 0.348 & 0.298 & 0.864 & 0.021 & 0.489 & 0.750\\
& & $S_{E}$ & -0.570 & 0.787 & 0.617 & -0.021 & 0.893 & 0.553\\
& & $S_{E}^*$ & \textbf{0.264} & \textbf{0.635} & \textbf{0.673} & \textbf{0.267} & \textbf{0.584} & \textbf{0.734}\\
\midrule
\multirow{7}{*}{regular\_plural\_subject\_verb\_agreement\_1} & & Random & 0.694 & 0.316 & 0.853 & 0.693 & 0.336 & 0.849\\
& & $S_{GN}$ & 0.740 & \textbf{0.194} & \textbf{0.946} & 0.724 & \textbf{0.268} & \textbf{0.906}\\
& & $S_{GN}^*$ & \textbf{0.756} & 0.251 & 0.909 & \textbf{0.747} & 0.274 & 0.898\\
&1.13 & $S_{GI}$ &  \textbf{0.748} & \textbf{0.202} & \textbf{0.944} & -0.039 & 0.333 & 0.852\\
& & $S_{GI}^*$ & 0.371 & 0.242 & 0.889 & \textbf{0.039} & \textbf{0.262} & \textbf{0.879}\\
& & $S_{E}$ & -0.614 & 0.610 & 0.718 & 0.303 & 0.632 & 0.694\\
& & $S_{E}^*$ & \textbf{0.584} & \textbf{0.353} & \textbf{0.836} & \textbf{0.568} & \textbf{0.313} & \textbf{0.842}\\
\bottomrule
\end{tabular}}
\caption{Alignment of GPT-2 and GPT-Neo explanations with BLiMP. Scores better than their (non-)contrastive counterparts are bolded. ``Dist'' gives the average distance from the target to the important context token.}
\label{table:align}
\end{table*}

\section{Highly Confusable Word Pairs}
\label{sec:pairs}
In Table \ref{table:pairs}, we provide the list of contrastive word pairs used in our human study for model simulatability (\S\ref{sec:human}). The first 10 pairs are taken from BLiMP linguistic paradigms and we provide the associated unique identifier for each pair. The last 10 pairs are chosen from word pairs with the highest confusion score.

\begin{table}[htp]
\centering
\resizebox{\linewidth}{!}{ 
\begin{tabular}{ccc}
\toprule
Word 1 & Word 2 & BLiMP UID \\
\midrule
actor & actress & anaphor\_gender\_agreement  \\
herself & himself & anaphor\_gender\_agreement \\
themselves & herself & anaphor\_number\_agreement \\
women & pictures & animate\_subject\_passive \\
boy & dog & animate\_subject\_passive \\
cat & cats & determiner\_noun\_agreement\_1 \\
is & are & irregular\_plural\_subject\_verb\_agreement\_1 \\
has & have & regular\_plural\_subject\_verb\_agreement\_1 \\
him & himself & principle\_A\_domain\_1 \\
he & who & wh\_island \\
\midrule
Word 1 & Word 2 & Confusion Score \\
\midrule
black & green & 0.0008 \\
Bruce & Beth & 0.0021 \\
fast & super & 0.0011 \\
health & hospital & 0.0012 \\
red & bright & 0.0007 \\
snow & winter & 0.0005 \\
son & brother & 0.0027 \\
summer & winter & 0.0003 \\
white & blue & 0.0034\\
wine & grape & 0.0106\\
\bottomrule
\end{tabular}}
\caption{List of highly confusable words pairs chosen for our user study.}
\label{table:pairs}
\end{table}

\section{Mixed Effects Models Results}
\label{sec:statsmodels}

In Table \ref{table:stats}, we show the results of fitting linear mixed-effects models to the results of our user study for model simulatability (\S\ref{sec:human}). 

\begin{table}[htp]
\centering
\resizebox{\linewidth}{!}{ 
\begin{tabular}{c|ccc}
\toprule
Dependent Variable & Intercept & Effect & P-Value \\
\midrule
Accuracy & 0.624 & 0.015 & 0.050\\
Acc. Correct & 0.744 & 0.026 & 0.005\\
Acc. Incorrect & 0.530 & -0.010 & 0.460\\
Useful & 0.570 & 0.063 & 0.000 \\
Acc. Useful & 0.677 & -0.020 & 0.513 \\
Acc. Useful & 0.450 & -0.009 & 0.444 \\
\bottomrule
\end{tabular}}
\caption{The dependent variables, intercepts, the effect of the explanation method on the dependent variable and its p-value in the linear mixed-effects models fitted to model simulatability results.}
\label{table:stats}
\end{table}

\section{Analysis of Foil Clusters}
\label{sec:analysis}

In Figure \ref{table:clusters}, we give a few examples of clusters and explanations we obtain for each part of speech. For each part of speech, we describe our findings in more detail in the following.

\paragraph{Adjectives.}
When the target word is an adjective, other foil adjectives that are semantically similar to the target are often clustered together. For example, when the target is \textit{``black''}, we find one cluster with various color adjectives, and we also find a different cluster with various adjectives relating to the race or nationality of a person.

We find that to distinguish between different adjectives, input words that are semantically close to the correct adjective are salient. For example to disambiguate the adjective \textit{``black''} from other colors, words such as \textit{``venom''} and \textit{``relativity''} are important.

\paragraph{Adpositions.}
When the target is an adposition, other adpositions are often in the same cluster.

To distinguish between different adpositions, the verb associated with the adposition is often useful to the LM. For example, when the target word is \textit{``from''}, verbs such as \textit{``garnered''} and \textit{``released''} helps the model distinguish the target from other adpositions that are less commonly paired with these verbs (e.g. \textit{``for'', ``of''}). As another example, for the target word \textit{``for''}, verbs that indicate a long-lasting action such as \textit{``continue''} and \textit{``lived''} help the model disambiguate.

\paragraph{Adverbs.}
When the target is an adverb, other adverbs are often clustered together. Sometimes, when the target is a specific type of adverb, such as an adverb of place, we can find a cluster with other adverbs of the same type.

Similarly to adpositions, LMs often use the verb associated with the target adverb to contrast it from other adverbs. For example, the verbs \textit{``dating''} and \textit{``traced''} are useful when the target is \textit{``back''}, and the verbs \textit{``torn''} and \textit{``lower''} are useful when the target is \textit{``down''}. 

\paragraph{Determiners.}
Other determiners are often clustered together when the target is a determiner. Particularly, when the target is a possessive determiner, we find clusters with other possessive determiners, and when the target is a demonstrative determiner, we find clusters with demonstrative determiners.

When the determiner is a gendered possessive determiner such as \textit{``his''}, proper nouns of the same gender, such as \textit{``John''} and \textit{``George''}, are often useful. For demonstrative determiners, such as \textit{``this''}, verbs that are usually associated with a targeted object, such as \textit{``achieve''} and \textit{``angered''} are useful. 

\paragraph{Nouns.}

When the target noun refers to a person, for example, \textit{``girl''}, foil nouns that also refer to a person form one cluster (e.g. \textit{``woman'', ``manager'', ``friend''}), commonly male proper nouns form another (e.g. \textit{``Jack'', ``Robin'', ``James''}), commonly female proper nouns form another (e.g. \textit{``Sarah'', ``Elizabeth'', ``Susan''}), and inanimate objects form a fourth (e.g. \textit{``window'', ``fruit'', ``box''}). 

When the target noun is an inanimate object, there are often two notable clusters: a cluster with singular inanimate nouns and a cluster with plural inanimate nouns. This suggests how clustering foils by explanations confirm that certain grammatical phenomena require similar evidence for disambiguation; in this case, determiner-noun agreement.

To predict a target animate noun such as \textit{``girl''} instead of foil nouns that refer to a non-female or older person, input words that are female names (e.g. \textit{``Meredith''}) or that refer to youth (e.g. \textit{``young''}) are useful. To disambiguate from male proper nouns, input words that refer to female people (e.g. \textit{``Veronica'', ``she''}) or adjectives related to the target (e.g. \textit{``tall''}) influence the model to generate a female common noun. To disambiguate from female proper nouns, adjectives and determiners are useful. To disambiguate from inanimate objects, words that describe a human or a human action (e.g. \textit{``delegate'', ``invented''}) are useful.

To predict a target inanimate noun such as \textit{``page''} instead of nouns that are also singular, input words with similar semantics are important such as \textit{``sheet'' and ``clicking''} are important. For plural noun foils, the determiner (e.g. ``a'') is important.

\paragraph{Numbers.}
When the target is a number, non-number words often form one cluster and other numbers form another cluster.

To disambiguate numbers from non-number words, input words related to enumeration or measurement are useful (e.g. ``age'', ``consists'', ``least''). To disambiguate words like \textit{``hundred''} and \textit{``thousand''} from other numbers such as \textit{``20''} or \textit{``five''}, input words used for counting (e.g. \textit{``two'', ``several''}) are useful, because \textit{``hundred''}s are countable in English.

\paragraph{Pronouns.}
When the target word is a gendered pronoun, foil pronouns of a different gender from the target form one cluster, foils with proper nouns of a different gender form a second cluster, and foils with proper nouns of the same gender as the target form a third cluster. This shows that the model uses similar evidence to make decisions to verify anaphor gender agreement. We also did not find foil clusters associated with distinguishing the number of the pronoun: often, these decisions follow directly from deciding between a pronoun and a proper noun, or deciding between a male and female pronoun. 

To disambiguate a gendered pronoun such as such as \textit{``he''}, from pronouns or proper nouns with different genders (e.g. \textit{``she''} or \textit{``Anna''}), proper nouns of the same gender as the target (e.g. \textit{``James''}) and other gendered pronouns or determiners (e.g. \textit{``his''}) are useful. To disambiguate from proper nouns of the same gender as the target, interestingly, the same proper noun as the foil appearing in the input is positively salient; GPT-2 is often influenced by previously appearing proper nouns to generate a pronoun instead. 

\paragraph{Verbs.}
When the target word is a verb, foil verbs that have a different verb form are often clustered together. This suggests that the model uses similar input features to verify subject-verb agreement.

When the target verb is in present participle form, auxiliary verbs in the input are useful (e.g. \textit{``is'', ``been''}) to distinguish from verbs in other forms. Similarly, when the target verb is in infinitive form, verbs in the same compound as the target verb are important.

\end{document}